\documentclass[a4paper,fleqn]{cas_ee}

\usepackage[numbers]{natbib}

\usepackage{graphicx} 

\usepackage[ruled,vlined]{algorithm2e}
\usepackage{amsmath}
\usepackage{array}
\usepackage{amssymb}
\usepackage{amsfonts}
\usepackage{booktabs}
\usepackage{multirow}
\usepackage{multicol}
\usepackage{mathrsfs}
\usepackage{amsthm}
\usepackage{xcolor}
\newtheorem{theorem}{Theorem}

\newcommand{\uwave}{\textcolor{red}}
\newcommand{\newcite}[1]{\citeauthor{#1}~(\citeyear{#1})}
\newcommand{\bfx}{\textbf{x}}
\newcommand{\bfz}{\textbf{z}}

\def\tsc#1{\csdef{#1}{\textsc{\lowercase{#1}}\xspace}}
\tsc{WGM}
\tsc{QE}
\tsc{EP}
\tsc{PMS}
\tsc{BEC}
\tsc{DE}

\begin{document}
%

\shorttitle{Rationalizing Predictions by Adversarial Information Calibration}
\title[mode = title]{Rationalizing Predictions by Adversarial Information Calibration\footnote{This article is a substantially revised and extended version a preliminary paper at AAAI 2021~\cite{sha2021learn}.}}


   
   \author[1,2,6]{Lei Sha}[orcid=0000-0001-5914-7590]

\cormark[1]
\ead{shalei@buaa.edu.cn}

\address[1]{Institute of Artificial Intelligence, Beihang University, China}
\address[2]{Department of Computer Science, University of Oxford, UK}
\address[3]{Department of Computer Science, University College London, UK}
\address[4]{Institute of Logic and Computation, TU Wien, Austria}
\address[5]{Alan Turing Institute, London, UK}

\address[6]{Zhongguancun Laboratory, Beijing, China}
\author[3]{Oana-Maria  Camburu}
\ead{o.camburu@cs.ucl.ac.uk}


\author[4,2,5]{Thomas Lukasiewicz}
\cormark[1]
\ead{thomas.lukasiewicz@cs.ox.ac.uk}

\cortext[cor1]{Corresponding author}


\begin{abstract}
Explaining the predictions of AI models is paramount in safety-critical applications, such as in legal or medical domains.
One form of explanation for a prediction is an extractive rationale, i.e., a subset of features of an instance that lead the model to give its prediction on that instance. For example, the subphrase ``he stole the mobile phone'' can be an extractive rationale for the prediction of ``Theft''.
Previous works on generating extractive rationales usually employ a two-phase model: a selector that selects the most important features (i.e., the rationale) followed by a predictor that makes the prediction based exclusively on the selected features. 
One disadvantage of these works is that the main signal for learning to select features comes from the comparison of the answers given by the predictor to the ground-truth answers. 
In this work, we propose to squeeze more information from the predictor via an information calibration method. More precisely, we train two models jointly: one is a typical neural model that solves the task at hand in an accurate but black-box manner, and the other is a selector-predictor model that additionally produces a rationale for its prediction. The first model is used as a guide for the second model. We use an adversarial technique to calibrate the information extracted by the two models such that the difference between them is an indicator of the missed or over-selected features. 
In addition, for natural language tasks, we propose a language-model-based regularizer to encourage the extraction of fluent rationales. 
Experimental results on a sentiment analysis task, a hate speech recognition task as well as on three tasks from the legal domain show the effectiveness of our approach to rationale extraction.

\end{abstract}


\maketitle

\section{Introduction}
Although deep neural networks have recently been contributing to state-of-the-art advances in various areas \cite{Krizhevsky:2012:ICD:2999134.2999257, hinton2012deep, DBLP:journals/corr/SutskeverVL14}, 
such models are often black-box, and therefore may not be deemed appropriate in situations where safety needs to be guaranteed, such as legal judgment prediction and medical diagnosis. Interpretable deep neural networks are a promising way to increase the reliability of neural models~\cite{sabour2017dynamic}. To this end, extractive rationales, i.e., subsets of features of instances on which models rely for their predictions on the instances, can be used as evidence for humans to decide whether to trust a prediction and more generally a~model. 

There are many different methods to explain a deep neural model, such as probing internal representations~\cite{hewitt-liang-2019-designing,conneau-etal-2018-cram,pimentel-etal-2020-information,voita-titov-2020-information,cifka-bojar-2018-bleu,vanmassenhove2017investigating}, adding interpretability to deep neural models~\cite{graves2014neural,sabour2017dynamic,grathwohl2019your,agarwal2020neural,chen2016infogan,sha2021multi}, and looking for global decision rules~\cite{holte1993very,cohen1995fast,letham2015interpretable,borgelt2005implementation,yang2017scalable,furnkranz2012foundations}. Extracting rationales belongs to the second category. 

Previous works use selector-predictor types of neural models to provide extractive rationales. More precisely, such models are composed of two modules: (i) a \textit{selector} that selects a subset of features of each input, and (ii) a  \textit{predictor} that makes a prediction based solely on the selected features. For example, \newcite{yoon2018invase} and \newcite{lei2016rationalizing} use a selector network to calculate a selection probability for each token in a sequence, then sample a set of tokens that is the only input of the predictor. Supervision is typically given only on the final prediction and not on the rationales. 
\newcite{paranjape-etal-2020-information} also uses information bottleneck to find a better trade-off between the sparsity and the final task performance. Note that gold rationale labels are required for semi-supervised training in \newcite{paranjape-etal-2020-information}.

An additional typical desideratum in natural language pro\-cessing (NLP) tasks is that the selected tokens form a semantically fluent rationale. To achieve this, \newcite{lei2016rationalizing} added a non-differential regularizer that encourages any two adjacent tokens to be simultaneously selected or unselected. The selector and predictor are jointly trained in a REINFORCE-style manner~\cite{williams1992simple}  because the sampling process and the regularizer are not differentiable.
\newcite{bastings2019interpretable} further improved the quality of the rationales by using a HardKuma regularizer that also encourages any two adjacent tokens to be selected or unselected together, which is differentiable and no need to use REINFORCE any more.

\begin{figure}[!t]
\begin{center}
\includegraphics[width=\linewidth]{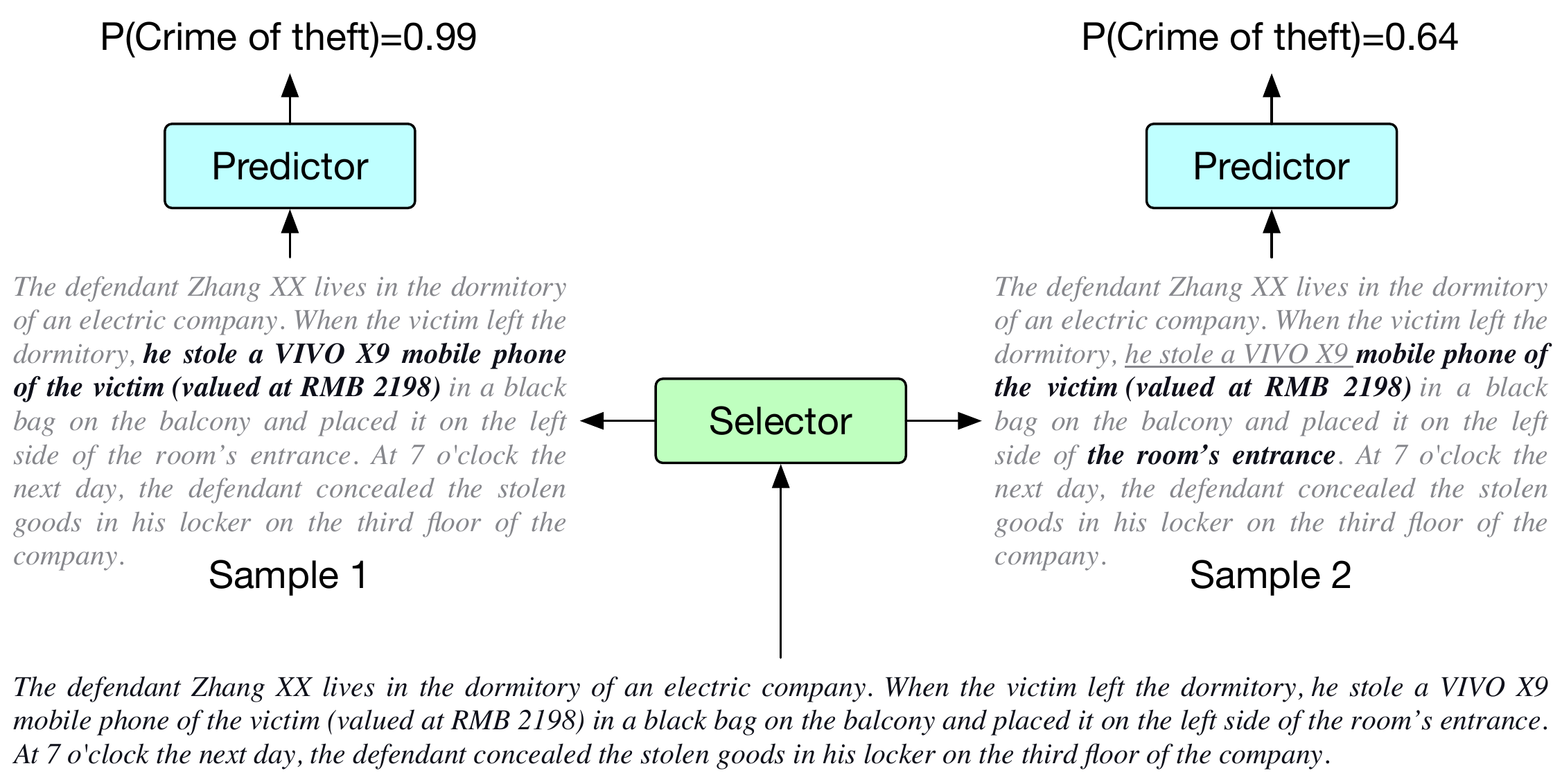}\\[0.75ex]
\caption{Examples of rationales in legal judgement prediction. The human-provided rationale is shown in bold in Sample~1. In Sample 2, the selector missed the key information ``he stole a VIVO X9'', but the predictor only tells the selector that the whole extracted rationale (in bold) is not so informative, by producing a low probability of the correct crime. }
\label{fig:intro}
\end{center}
\end{figure}

One drawback of previous works is that the learning signal for both the selector and the predictor comes mainly from comparing the prediction of the selector-predictor model with the ground-truth answer. 
Therefore, the exploration space to get to the correct rationale is large, decreasing the chances of converging to the optimal rationales and predictions. 
Moreover, in NLP applications, the regularizers commonly used for achieving fluency of rationales treat all adjacent token pairs in the same way. This often leads to the selection of unnecessary tokens due to their adjacency to informative~ones. 

In this work, we first propose an alternative method to rationalize the predictions of a neural model. Our method aims to squeeze more information from the predictor in order to guide the selector in selecting the rationales. Our method trains two models jointly: a ``guider'' model that solves the task at hand in an accurate but black-box manner, and a selector-predictor model that solves the task while also providing rationales. We use an adversarial-based training procedure to encourage the final information vectors generated by the two models to encode the same information. We use an information bottleneck technique in two places: (i)~to encourage the features selected by the selector to be the least-but-enough features, and (ii)~to encourage the final information vector of the guider model to also contain the least-but-enough information for the prediction.   
Secondly, we propose using language models as regularizers for rationales in natural language understanding tasks. A language model (LM) regularizer encourages rationales to be fluent subphrases, which means that the rationales are formed by consecutive tokens while avoiding unnecessary tokens to be selected simply due to their adjacency to informative tokens.  
The effectiveness of our LM-based regularizer is proved both by a mathematical derivation and experiments. 

The contributions of this article are briefly summarized as follows:
\begin{itemize}
\item We introduce a novel model that generates extractive rationales for its predictions. The model is based on an adversarial approach that calibrates the information between a guider and a selector-predictor model, such that the selector-predictor model learns to mimic a typical neural model while additionally providing rationales.


\item We propose a language-model-based regularizer to encourage the sampled tokens to form fluent rationales. Usually, this regularizer will encourage fewer fragment of subsequences and avoid strange start and end of the sequences. This regularizer  also gives priority to important adjacent token pairs, which benefits the extraction of informative features.

\item We experimentally evaluate our method on a sentiment analysis dataset and a hate speech detection dataset, both containing ground-truth rationale annotations for the ground-truth labels, as well as on three tasks of a legal judgement prediction dataset, for which we conducted human evaluations of the extracted rationales. 
The results show that our method improves over the previous state-of-the-art models in precision and recall of rationale extraction without sacrificing the prediction performance.
\end{itemize}


The rest of this paper is organized as follows. In Section~\ref{sec:approach}, we introduce our proposed approach, including the selector-predictor module (Section~\ref{sec:sp}), the guider module (Section~\ref{sec:guider}), the information calibrating method (Section~\ref{sec:cali}), and the language model-based rationale regularizer (Section~\ref{sec:LM}). In Section~\ref{sec:experiment}, we report the experimental results on the three datasets: a beer review dataset (Section~\ref{sec:beer}), a legal judgment prediction dataset (Section~\ref{sec:law}), and a hate speech detection dataset (Section~\ref{sec:hate}). Section~\ref{sec:related} reviews the related works of this paper. In Section~\ref{sec:conclu}, we provide a summary and an outlook on future research. 

\section{Related Work}\label{sec:related}
Explainability is currently a key bottleneck of deep-lear\-ning-based approaches~\cite{atkinson2020explanation,kaptein2021evaluating}. A summarization of related works  is shown in Table~\ref{tab:rel}, where we have listed the representative works in each branch of interpretable models. As is shown, previous works on explainable neural models include self-explanatory models and post-hoc explainers. 
The model proposed in this work belongs to the class of self-explanatory models, which contain an explainable structure in the model architecture, thus providing explanations / rationales for their predictions. Self-explanatory models can use different types of explanations / rationales, such as feature-based explanations which is usually conducted by selector-predictors~ \cite{lei2016rationalizing, yoon2018invase,chen2018learning,yu-etal-2019-rethinking,carton-etal-2018-extractive} and natural language explanations 
\cite{DBLP:conf/eccv/HendricksARDSD16, esnli, zeynep, cars}. Our model uses feature-based explanations.

\begin{table}[]
\renewcommand{\arraystretch}{1.4}
\resizebox{\linewidth}{!}{
\begin{tabular}{|l|l|p{3cm}|p{5cm}|p{1.5cm}|p{1.5cm}|p{1.5cm}|p{1.5cm}|p{1.5cm}|}
\hline
 & & & Representative Methods & Controllable & Provide important features& Provide important examples & Provide NL explanations & Provide rules\\
\hline
\multirow{12}{1.5cm}{Self-explanatory methods} & \multirow{2}{2cm}{Disentangle\-ment} & Implicit                          &$\beta$-VAE~\cite{higgins2017beta}, $\beta$-TCVAE~\cite{chen2018isolating} & Yes & && &\\ \cline{3-9} 
                                           &    & Explicit  & InfoGAN~\cite{chen2016infogan}, MTDNA~\cite{sha2021multi} & Yes& Yes && &   \\ \cline{2-9}
                                           & \multirow{11}{*}{Architecture}    & \multirow{2}{*}{Attention-based}  & \newcite{rocktaschel2015reasoning}, \newcite{vaswani2017attention}, OrderGen~\cite{sha2018order} &  & Yes && & \\ \cline{3-9} 
                                           &                                  & \multirow{3}{*}{Read-Write Memory}    & Neural Turing Machines~\cite{collier2018im,sha2020estimating}, Progressive Memory~\cite{rusu2016progressive,xia-etal-2017-progressive}, Differentiable neural computer~\cite{graves2016hybrid}, Neural RAM~\cite{kaiser2015neural}, Neural GPU~\cite{kurach2015neural} &Yes & && & \\ \cline{3-9}
                                           &                                  & Capsule-based  & Capsule~\cite{sabour2017dynamic} &  & Yes && & \\ \cline{3-9} 
                                           &                                  & \multirow{3}{*}{Energy-based} & \newcite{grathwohl2019your}, Hopfield Network~\cite{ramsauer2020hopfield}, Boltzmann Machine~\cite{marullo2021boltzmann}, Predictive Coding~\cite{song2020can} &  & Yes && & \\\cline{3-9} &   & \multirow{2}{*}{Rationalization} & Selector-predictor~\cite{lei2016rationalizing,bastings2019interpretable,yoon2018invase,chen2018learning}, natural language explanations~\cite{DBLP:conf/eccv/HendricksARDSD16,esnli} &  & Yes & &Yes& \\    \hline
\multirow{14}{1.5cm}{Post-hoc explainer}       & \multirow{8}{*}{Local}          & \multirow{1}{*}{Perturbation-based}               & SHAP~\cite{lundberg2017unified}, Shapley Values~\cite{shapley1953value} &  & Yes & &&\\ \cline{3-9} 
                                           &                                  & \multirow{1}{*}{Surrogate-based}   &Anchors~\cite{ribeiro2018anchors}, LIME~\cite{ribeiro2016should} & & Yes&& & Yes   \\ \cline{3-9} 
                                           &                                  & \multirow{2}{*}{Saliency Maps}    &Input gradient~\cite{baehrens2010explain,simonyan2014deep,shrikumar2017learning}, SmoothGrad~\cite{smilkov2017smoothgrad,seo2018noise}, Integrated Gradients~\cite{sundararajan2017axiomatic}, Guided Backprop~\cite{springenberg2015striving}  & & Yes & & &  \\ \cline{3-9} 
                                           &                                  & \multirow{2}{3cm}{Prototypes/Example Based}          &Influence Functions~\cite{cook1980characterizations,koh2017understanding}, Representer Points~\cite{yeh2018representer}, TracIn~\cite{pruthi2020estimating}  & & &Yes & & \\ \cline{3-9} 
                                           &                                  & \multirow{2}{*}{Counterfactuals}  &\newcite{wachter2017counterfactual}, \newcite{mahajan2019preserving}, \newcite{karimi2020algorithmic}  & & Yes & Yes & & \\ \cline{2-9} 
                                           & \multirow{5}{*}{Global}          & Collection of Local Explanations &SP-LIME~\cite{ribeiro2016should}, Summaries of Counterfactuals~\cite{rawal2020beyond}  & &Yes & & &  \\\cline{3-9} 
                                           &                                  & \multirow{2}{*}{Model Distillation}  & Tree Distillation~\cite{bastani2017interpreting}, Decision set distillation~\cite{lakkaraju2019faithful}, Generalized Additive Models~\cite{tan2018learning} & & Yes & & & Yes  \\ \cline{3-9} 
                                           &                                  & Representation based &Network Dissection~\cite{bau2017netdissect}, TCAV~\cite{kim2018interpretability}  & & Yes & & & \\ \hline
\end{tabular}
}
\caption{The branches of related works.}
\label{tab:rel}
\end{table}

\subsection{Self-explanatory models for interpretability.}
Self-explanatory models with feature-based explanations can be further divided into two branches. The first branch is disentanglement-based approaches, which map specific features into latent spaces and then use the latent variables to control~the outcomes of the model, such as disentangling methods~\cite{chen2016infogan,sha2021multi}, information bottleneck methods~\cite{tishby2000information}, and constrained generation~\cite{sha-2020-gradient}. The second branch consists of architecture-in\-ter\-pretable models, such as attention-based models \cite{zhang2018top,sha2016reading,sha2018order,sha2018multi,liu2017table}, Neural Turing Machines~\cite{collier2018im,xia-etal-2017-progressive,sha2020estimating}, capsule networks~\cite{sabour2017dynamic}, and energy-based models~\cite{grathwohl2019your}. Among them, attention-based models have an important extension, that of sparse feature learning, which implies learning to extract a subset of features that are most informative for each example. Most of the sparse feature learning methods use a selector-predictor architecture. Among them, L2X \cite{chen2018learning} and INVASE~\cite{yoon2018invase} make use of information theories for feature selection, while CAR~\cite{chang2019game} extracts useful features in a game-theoretic approach.

In addition, rationale extraction for NLP usually raises one desideratum for the extracted subset of tokens: rationales need to be fluent subphrases instead of separate tokens. To this end, \newcite{lei2016rationalizing} proposed a non-differentiable regularizer to encourage selected tokens to be consecutive, which can be optimized by REINFORCE-style methods~\cite{williams1992simple}. \newcite{bastings2019interpretable} proposed a differentiable regularizer using the Hard Kumaraswamy distribution; however, this still does not consider the difference in the importance of different adjacent token pairs. \newcite{paranjape-etal-2020-information} proposed a very similar information bottleneck method to our InfoCal method. However, they did not use any calibration method to encourage the completeness of  the extracted rationale. 

Our method belongs to the class of self-explanatory methods. Different from previous sparse feature learning methods, we use an adversarial information calibrating mechanism to hint to the selector about missing important features or over-selected features. Moreover, our proposed LM regularizer is differentiable and can be directly optimized by gradient descent. This regularizer also encourages important adjacent token pairs to be simultaneously selected, which benefits the extraction of useful features. 

\subsection{Post-hoc explainers  for interpretability}
Post-hoc explainers analyze the effect of each feature in the prediction of an already trained and fixed model. Post-hoc explainers can be divided into two types: local explainers and global explainers. Local explainers can be further split into five categories: (a)  perturbation-based:    change the values of some features to see their effect on the outcome~\cite{friedman2001greedy,hooker2004discovering,friedman2008predictive,fisher2018model,greenwell2018simple,zhao2019causal,goldstein2015peeking,janzing2019feature,apley2020visualizing}. Some famous  perturbation-based post-hoc explainer methods include Shapley values~\cite{shapley1953value,vstrumbelj2014explaining,sundararajan2020many,janzing2020feature,staniak2018explanations} and SHAP method~\cite{lundberg2017unified,lundberg2018consistent,slack2020fooling}, (b) surrogate-based: train an explainable model, such as linear regression or decision trees, to approximate the predictions of a black-box model~\cite{kaufmann2013information,alvarez2018robustness,ribeiro2018anchors,slack2020fooling}, for example, LIME~\cite{ribeiro2016should}. (c) saliency maps: use gradient information to show what parts of the input are most relevant for the model's prediction, including input gradient~\cite{baehrens2010explain,simonyan2014deep,shrikumar2017learning}, SmoothGrad~\cite{smilkov2017smoothgrad,seo2018noise}, integrated gradients~\cite{sundararajan2017axiomatic}, guided backprop~\cite{springenberg2015striving}, class activation mapping~\cite{zhou2016learning}, meaningful perturbation~\cite{fong2017interpretable}, RISE~\cite{petsiuk2018rise}, extremal perturbations~\cite{fong2019understanding}, DeepLift~\cite{shrikumar2017learning}, expected gradients~\cite{erion2019improving}, excitation backprop~\cite{zhang2018top}, GradCAM~\cite{selvaraju2017grad}, occlusion~\cite{zeiler2014visualizing}, prediction difference analysis~\cite{gu2019contextual}, and internal influence~\cite{leino2018influence}. (d) prototype / example based: find which training example affects the model prediction the most. Usually, this is conducted by influence functions~\cite{cook1980characterizations}. (e) counterfactual explanations: detect what features need to be changed to flip the model's prediction~\cite{wachter2017counterfactual,mahajan2019preserving,karimi2020algorithmic}. On the other hand, some of the global explainers are   collections of local explanations (e.g., SP-LIME~\cite{ribeiro2016should}, and summaries of counterfactuals~\cite{rawal2020beyond}). Also, distillation methods provide explainable rules by distilling the information from deep models to tree models~\cite{bastani2017interpreting} or decision set models~\cite{lakkaraju2019faithful}. There are also some methods~(Network Dissection~\cite{bau2017netdissect}, TCAV~\cite{kim2018interpretability}) derives model understanding by analyzing intermediate representations of a deep black-box model.

\subsection{Information bottleneck}
The information bottleneck~(IB) theory is an important basic theory of neural networks ~\cite{tishby2000information}. It originated in information theory and has been widely used as a theoretical framework in analyzing deep neural networks~\cite{tishby2015deep}. For example, \newcite{li-eisner-2019-specializing} used IB to compress word embeddings in order to make them contain only specialized information, which leads to a much better performance in parsing tasks.

\subsection{Adversarial methods}
Adversarial methods, which had been widely applied in image generation~\cite{chen2016infogan} and text generation~\cite{yu2017seqgan}, usually have a discriminator and a generator. The discriminator receives pairs of instances from the real distribution and from the distribution generated by the generator, and it is trained to differentiate between the two. The generator is trained to fool the discriminator~\cite{goodfellow2014generative}. Our information calibration method generates a dense feature vector using selected symbolic features, and the discriminator is used for measuring the calibration extent. 

Our adversarial calibration method is inspired by distilling methods~\cite{hinton2015distilling}. 
Distilling methods are usually applied to compress large models into small models while keeping a comparable performance. For example, TinyBERT~\cite{jiao2019tinybert} is a distillation of BERT~\cite{devlin-etal-2019-bert}. Our method is different from distilling methods, because we calibrate the final feature vector instead of the softmax prediction. Also, to our best knowledge, we are the first to apply information calibration for rationale extraction.

\section{Approach}\label{sec:approach}
Our approach is composed of a selector-predictor architecture, in which  we use the information bottleneck technique to restrict the number of selected features, and a guider model, for which we again use the information bottleneck technique to restrict the information in the final feature vector. Then, we use an adversarial method to make the guider model guide the selector into selecting the least-but-enough features. Finally, we use a language model (LM) regularizer to obtain semantically fluent rationales.

\subsection{InfoCal: Selector-Predictor-Guider with Information Bottleneck}\label{sec:sp}

The Selector-Predictor-Guider architecture contains two parallel architectures, one is a selector-predictor model, which selects the rationale and judges whether it can make a correct prediction; the other is a guider model, which is a dense ``black-box'' neural network trying to learn the feature vector required for the task. The information calibration is used to calibrate the dense feature vector learned by the guider model and the information contained in the rationales extracted by the selector-predictor model.
The high-level architecture of our model, called InfoCal, is shown in Fig.~\ref{fig:arch}. 
Below, we detail each of its components.  

\subsubsection{Selector}

For a given instance $(\bfx,y)$, $\bfx$ is the input with $n$ features $\bfx = (x_1, x_2, \ldots, x_n)$, and $y$ is the ground-truth corresponding label. The selector network $\text{Sel}(\tilde{\bfz}_\text{sym}|\bfx)$ takes $\bfx$ as input and outputs $p(\tilde{\bfz}_\text{sym}|\bfx)$, a sequence of probabilities $(p_i)_{i=1,\ldots,n}$ representing the probability of choosing each feature $x_i$ as part of the rationale. 

Given the sampling probabilities, a subset of features is sampled using the Gumbel softmax~\cite{jang2016categorical}, which provides a differentiable sampling process: 
\begin{align}
u_i&\sim U(0,1),\quad g_i=-\log(-\log(u_i))\\
m_i&=\frac{\exp((\log(p_i)+g_i)/\tau)}{\sum_j\exp((\log(p_j)+g_j)/\tau)},\label{eq:maski}
\end{align}
where $U(0,1)$ represents the uniform distribution between $0$ and $1$, and $\tau$ is a temperature hyperparameter. 
Hence, we obtain the sampled mask $m_i$ for each feature $x_i$, and the vector symbolizing the rationale $\tilde{\bfz}_\text{sym}=(m_1x_1,\ldots,m_nx_n)$. Thus, $\tilde{\bfz}_\text{sym}$ is the sequence of discrete selected symbolic features forming the rationale.

\subsubsection{Predictor} The predictor takes as input the rationale $\tilde{\bfz}_\text{sym}$ given by the selector, and outputs the prediction $\hat{y}_{sp}$. 
In the selector-predictor part of InfoCal, the input to the predictor is the multiplication of each feature $x_i$ with the sampled mask $m_i$. The predictor first calculates a dense feature vector $\tilde{\bfz}_\text{nero}$,\footnote{Here, ``nero'' stands for neural feature (i.e., a neural vector representation) as opposed to a symbolic input feature.} then uses one feed-forward layer and a softmax layer to calculate the probability distribution over the possible predictions: 
\begin{align}
\tilde{\bfz}_\text{nero}&=\text{Pred}(\tilde{\bfz}_\text{sym})\\
p(\hat{y}_{sp}|\tilde{\bfz}_\text{sym}) &= \text{Softmax}(W_p\tilde{\bfz}_\text{nero}+b_p).\label{eq:pyn}
\end{align}
As the input is masked by $m_i$, the prediction $\hat{y}_{sp}$ is exclusively based on the features selected by the selector. The loss of the selector-predictor model is the cross-entropy loss:
\begin{equation}\label{eq:lsp}
\begin{small}
\begin{aligned}
L_{sp}&=-\frac{1}{K}\sum_k\log p( y^{(k)}_\text{sp}|\bfx^{(k)})\\
&=-\frac{1}{K}\sum_k\log \mathbb E_{\text{Sel}(\tilde{\bfz}_\text{sym}^{(k)}|\bfx^{(k)})}p(y^{(k)}_\text{sp}|\tilde{\bfz}_\text{sym}^{(k)})\\
&\le -\frac{1}{K}\sum_k\mathbb E_{\text{Sel}(\tilde{\bfz}_\text{sym}^{(k)}|\bfx^{(k)})}\log p(y^{(k)}_\text{sp}|\tilde{\bfz}_\text{sym}^{(k)}), 
\end{aligned}
\end{small}
\end{equation}
where $K$ represents the size of the training set, the superscript (k) denotes the k-th instance in the training set, and the inequality follows from Jensen's inequality.

\subsubsection{Guider} \label{sec:guider}
\begin{figure}[!t]
\begin{center}
\includegraphics[width=0.8\linewidth]{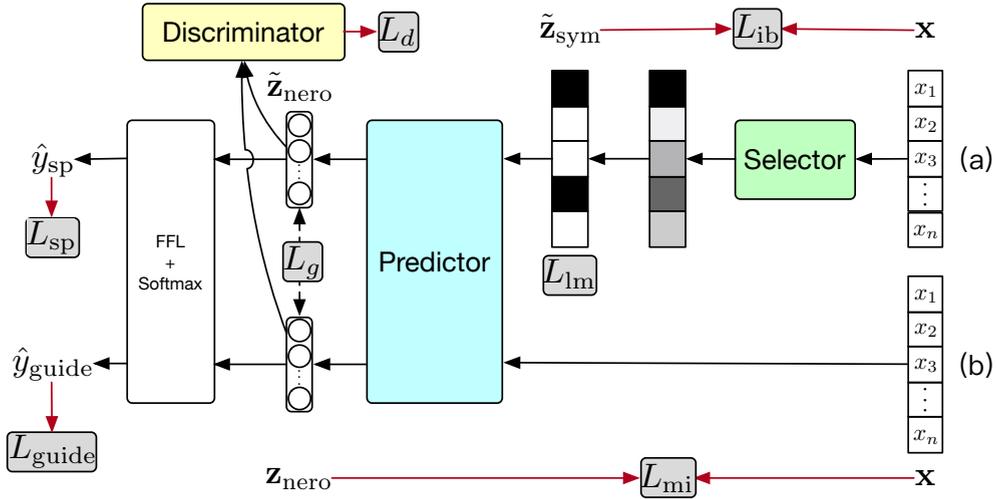}\\
\caption{Architecture of InfoCal: the grey round boxes stand for the losses, and the red arrows indicate the data required for the calculation of the losses. FFL is an abbreviation for feed-forward layer.}
\label{fig:arch}
\end{center}
\end{figure}

To guide the rationale selection of the selector-predictor model, we train a \textit{guider} model, denoted Pred$_G$, which receives the full original input $\bfx$ and transforms it into a dense feature vector $\bfz_\text{nero}$, using the same predictor architecture as the selector-predictor module, but different weights, as shown in Fig.~\ref{fig:arch}. We generate the dense feature vector in a variational way, which means that we first generate a Gaussian distribution according to the input $\bfx$, from which we sample a vector $\bfz_\text{nero}$: 
\begin{align}
h&=\text{Pred}_G(\bfx),\quad\mu=W_mh+b_m,\quad\sigma=W_sh+b_s \\
u&\sim \mathcal N(0,1), \quad \bfz_\text{nero} = u\sigma + \mu\\
p&(\hat y_\text{guide}|\bfz_\text{nero}) = \text{Softmax}(W_p\bfz_\text{nero}+b_p).
\end{align}
We use the reparameterization trick of Gaussian distributions to make the sampling process differentiable~\cite{kingma2013auto}. We share the parameters $W_p$ and $b_p$ with those in Eq.~\ref{eq:pyn}.

The guider model's loss $L_\text{guide}$ is as follows:
\begin{equation}\label{eq:full}
\begin{aligned}
L_\text{guide}&=-\frac{1}{K}\sum_k\log p(y^{(k)}_\text{guide}|\bfx^{(k)})\\
&\le-\frac{1}{K}\sum_k\mathbb E_{p(\bfz_\text{nero}|\bfx^{(k)})}\log p(y^{(k)}_\text{guide}|\bfz_\text{nero}^{(k)}), 
\end{aligned}
\end{equation}
where the inequality again follows from Jensen's inequality. The guider and the selector-predictor are trained jointly.

\subsubsection{Information Bottleneck} \label{sec:ib}
To guide the model to select the least-but-enough information, we employ an information bottleneck technique \cite{li-eisner-2019-specializing}. We aim to minimize $I(\bfx, \tilde{\bfz}_\text{sym}) - I(\tilde{\bfz}_\text{sym},y)$\footnote{$I(a,b) = \int_a\int_bp(a,b)\log\frac{p(a,b)}{p(a)p(b)} \,{=}\, \mathbb E_{a,b}[\frac{p(a|b)}{p(a)}]$ denotes  the mutual information between the variables $a$ and  $b$.}, where the former term encourages the selection of few features, and the latter term encourages the selection of the necessary features.  As $I(\tilde{\bfz}_\text{sym},y)$ is implemented by $L_{sp}$ (the proof is given in Appendix \ref{the:0}), we only need to minimize the mutual information $I(\bfx, \tilde{\bfz}_\text{sym})$:
\begin{align}\label{eq:Isym}
I(\bfx, \tilde{\bfz}_\text{sym})=\mathbb E_{\bfx, \tilde{\bfz}_\text{sym}}\Big[\log\frac{p(\tilde{\bfz}_\text{sym}|\bfx)}{p(\tilde{\bfz}_\text{sym})}\Big].
\end{align}

However, there is a time-consuming term $p(\tilde{\bfz}_\text{sym})=\sum_{\bfx}p(\tilde{\bfz}_\text{sym}|\bfx)p(\bfx)$, which needs to be calculated by a loop over all the instances $\bfx$ in the training set. Inspired by \newcite{li-eisner-2019-specializing}, we replace this term with a variational distribution $r_\phi(z)$ and obtain an upper bound of Eq.~\ref{eq:Isym}: $I(\bfx, \tilde{\bfz}_\text{sym}) \le \mathbb E_{\bfx, \tilde{\bfz}_\text{sym}}\Big[\log\frac{p(\tilde{\bfz}_\text{sym}|\bfx)}{r_\phi(z)}\Big]$. Since $\tilde{\bfz}_\text{sym}$ is a sequence of binary-selected features, we sum up the mutual information term of each element of $\tilde{\bfz}_\text{sym}$ as the information bottleneck loss:
\begin{align}
L_\text{ib}=\sum_i\sum_{\tilde{z}_i} p(\tilde{z}_i|\bfx)\log\frac{p(\tilde{z}_i|\bfx)}{r_\phi(z_i)}, 
\end{align}
where $\tilde{z}_i$ represents whether to select the $i$-th feature: $1$ for selected, $0$ for not selected.

To encourage $\bfz_\text{nero}$ to contain the least-but-enough information in the guider model, we again use the information bottleneck technique. 
Here, we minimize $I(\bfx, \bfz_\text{nero}) - I(\bfz_\text{nero},y)$. Again, $I(\bfz_\text{nero},y)$ can be implemented by $L_\text{guide}$.
Due to the fact that $\bfz_\text{nero}$ is sampled from a Gaussian distribution, the mutual information has a closed-form upper bound:
\begin{equation}\label{eq:mi}
\begin{aligned}
L_\text{mi}&=I(\bfx, \bfz_\text{nero})\le \mathbb E_{\bfz_\text{nero}}\Big[\log\frac{p(\bfz_\text{nero}|\bfx)}{p(\bfz_\text{nero})}\Big] =0.5(\mu^2+\sigma^2-1-2\log\sigma).
\end{aligned}
\end{equation}
The derivation is in Appendix~\ref{proof:mi}.

\subsection{Calibrating Key Features via Adversarial Training}\label{sec:cali}
Our goal is to inform the selector what kind of information is still missing or has been wrongly selected. Since we already use the information bottleneck principal to encourage $\bfz_\text{nero}$ to encode the information from the least-but-enough features, if we also require $\tilde{\bfz}_\text{nero}$ and $\bfz_\text{nero}$ to encode the same information, then we would encourage the selector to select the least-but-enough discrete features. 
To achieve this, we use an adversarial-based training method.
Thus, we employ an additional discriminator neural module, called~$D$, which takes as input either $\tilde{\bfz}_\text{nero}$ or $\bfz_\text{nero}$ and outputs label ``0'' or label ``1'', respectively. The discriminator can be any differentiable neural network. The generator in our model is formed by the selector-predictor that outputs $\tilde{\bfz}_\text{nero}$. 
The losses associated with the generator and discriminator are:
\begin{align}
L_d&=-\log D(\bfz_\text{nero}) + \log D(\tilde{\bfz}_\text{nero})\label{eq:D}\\
L_g&=- \log D(\tilde{\bfz}_\text{nero}).
\end{align}

\citet{yoon2018invase} also attempted to use guidance from a so-called ``base" model to a selector-predictor model. 
Nevertheless, their ``base'' model can only provide valid information calibration in actor-critic reinforcement learning, which is difficult to provide in POMDP problems~\cite{kaelbling1998planning}. In comparison, the discriminator in our method is more flexible in providing valid information calibration.

\subsection{Regularizing Rationales with Language Models}
\label{sec:LM}

For NLP tasks, it is often desirable that a rationale is formed of 
fluent subphrases
\cite{lei2016rationalizing}. To this end, previous works propose regularizers that bind the adjacent tokens to make them be simultaneously sampled or not. For example, \newcite{lei2016rationalizing} proposed a non-differentiable regularizer trained using REINFORCE~\cite{williams1992simple}. To make the method differentiable, \newcite{bastings2019interpretable} used the Kumaraswamy-distribution for the regularizer.
However, they treat all pairs of adjacent tokens in the same way, even though some adjacent tokens have more priority to be bound than others, such as  ``He stole'' or ``the victim'' rather than ``. He'' or ``) in'' in Fig.~\ref{fig:intro}. 


We propose a novel differentiable regularizer for extractive rationales that is based on a pre-trained language model, thus encouraging both the consecutiveness and the fluency of the tokens in the extracted rationale. 
The LM-based regularizer is implemented as follows:
\begin{equation}\label{eq:lm}
    L_\text{lm} = -\sum_im_{i-1}\log p_{lm}(m_ix_i|\bfx_{<i}),
\end{equation}
where the $m_i$'s are the masks obtained in Eq.~\ref{eq:maski}. 
Note that non-selected tokens are masked instead of deleted in this regularizer. The language model 
can have any architecture. 

First, we note that $L_\text{lm}$ is differentiable. Second, the following theorem guarantees that $L_\text{lm}$ encourages consecutiveness of selected tokens.

\begin{theorem}\label{the:1}
If the following is satisfied for all $i,j$: 
\begin{itemize}
    \item $m'_i<\epsilon \ll 1-\epsilon< m_i$, \,$0<\epsilon<1$, and 
    \item $\big|p(m'_ix_i|x_{<i})-p(m'_jx_j|x_{<j})\big|<\epsilon$,
\end{itemize}
then the following two inequalities hold:\\
 (1) $L_\text{lm}(\ldots,m_k,\ldots, m'_{n})<L_\text{lm}(\ldots,m'_k,\ldots, m_{n})$.\\
 (2) $L_\text{lm}(m_1, \ldots,m'_k,\ldots)>L_\text{lm}(m'_1,\ldots,m_k,\ldots)$.
\end{theorem}
The theorem says that for the same number of selected tokens, if they are consecutive, then they will get a lower $L_\text{lm}$ value.
Its proof is given in Appendix \ref{proof:1}. 

\subsubsection{Language Model in Continuous Form} \label{lmvec}
 
Conventional language models are in discrete-form, which usually generate a multinomial distribution for each token, and minimize the Negative Log-likelihood  (NLL) loss. The probability of the expected token is computed as follows:
 \begin{equation}\label{eq:lm_lastlayer}
p(x_i|x_{<i})=\frac{\exp{(h_i^\top e_i)}}{\sum_{j\in\mathcal V}\exp{(h_i^\top e_j)}},
\end{equation}
where $h_i$ is the hidden vector corresponding to $x_i$, $e_i$ is a trainable parameter which represents the output vector of $x_i$, and $\mathcal V$ is the vocabulary. In language model literature~\cite{kuhn1990cache,bengio2003neural}, $x_i$ is a symbolic token, and each token in $\mathcal V$ has a corresponding trainable output vector. Eqn.~\ref{eq:lm_lastlayer} is a \texttt{Softmax} operation which normalizes throughout the whole vocabulary.

Note that in Eq.~\ref{eq:lm}, the target sequence of the language model $P(m_ix_i|x_{<i})$ is formed of vectors instead of symbolic tokens. Since $m_ix_i$ is not symbolic token, it do not have a corresponding trainable output vector so that we cannot use a \texttt{Softmax}-like operation to normalize throughout the whole vocabulary. To tackle  this, we require a continuous-form language model. Therefore, we  make some small changes in the pre-training of the language model. 
 When we are modeling the language model in vector form, we only use a bilinear layer to directly calculate the probability in Eq.~\ref{eq:lm_lastlayer}: 
 \begin{equation}\label{eq:vlm}
 p(x_i|x_{<i})=\sigma(h_i^\top Me_i),
 \end{equation}
  where $\sigma$ stands for \texttt{sigmoid}, and $M$ is a trainable parameter matrix. The \texttt{sigmoid} operation ensures the result lies in $[0,1]$, which is a probability value. Then the probability value of $P(m_ix_i|x_{<i})$ is computed by:
   \begin{equation}
 p(m_ix_i|x_{<i})=\sigma(h_i^\top M(m_ie_i)).
 \end{equation}
 
  However, without normalization operations like \texttt{Softmax}, what Eqn.~\ref{eq:vlm} computes is a quasi-probability value which   relates to only one token. 
  To solve this issue, we use negative sampling~\cite{mikolov2013distributed} in the training procedure. Therefore, the language model is pretrained using the following loss:
\begin{equation}\label{eq:pre}
 L_\text{pre}=-\sum_i\Big[\log\sigma(h_i^\top Me_i) - \mathbb E_{j\sim p(x_j)}\log\sigma(h_i^\top Me_j) \Big], 
\end{equation}
where $p(x_j)$ is the occurring probability (in the training dataset) of token $x_j$.

\subsection{Training and Inference}
The total loss function of our model, which takes the generator's role in adversarial training, is shown in Eq.~\ref{eq:G}. The adversarial-related losses are denoted by $L_\text{adv}$. The discriminator is trained by $L_d$ from Eq.~\ref{eq:D}.
\begin{align}
L_\text{adv} &= \lambda_{g}L_g + L_{guide} + \lambda_{mi}L_\text{mi}\\
J_\text{total}&=L_{sp} +\lambda_{ib}L_\text{ib}+L_\text{adv}+\lambda_{lm}L_\text{lm}, \label{eq:G}
\end{align}
where $\lambda_{ib},\lambda_{g},\lambda_{mi}$, and $\lambda_{lm}$ are hyperparameters.

At training time, we optimize the generator loss $J_\text{total}$ and discriminator loss $L_d$ alternately until convergence. 
At inference time, we run the selector-predictor model to obtain the prediction and the rationale~$\tilde{\bfz}_\text{sym}$.

The whole training process is illustrated in Algorithm~\ref{algo:1}.

\begin{algorithm}
\SetAlgoLined
 Random initialization\;
 Pre-train language model by Eq.~\ref{eq:pre}\;
 \For{each iteration $i=1,2,\ldots$ }{
 \For{each batch  }{
  Calculate the loss $J_\text{total}$ for the sampler-predictor model and the guider model by Eq.~\ref{eq:G}\;
  Calculate the loss $L_D$ for the discriminator by Eq.~\ref{eq:D}\;
  Update the parameters of selector-predictor model and the guider model\;
  Update the parameters of the discriminator\;
  }
 }
 \caption{Training process of InfoCal.}
\label{algo:1}
\end{algorithm}

\section{Experiments}\label{sec:experiment}

We performed experiments on three NLP applications: multi-aspect sentiment analysis, legal judgement prediction, and hate speech detection.
For multi-aspect sentiment analysis and hate speech detection, we have rationale annotations in the dataset. So, we can directly use automatic evaluation metrics to evaluate the quality of extracted rationales. For legal judgement prediction, there is no rationale annotation, so we conduct human evaluation for the extracted rationales.

\subsection{Evaluation Metrics for Rationales.}
With the annotations of rationales in the multi-aspect sentiment analysis and hate speech detection datasets, we would like to evaluate the  explainability of our model. For better comparison, we use the same evaluation metrics  with previous works~\cite{deyoung-etal-2020-eraser,mathew2020hatexplain}, which contains 5 metrics  as listed below. 
\begin{itemize} 
    \item IOU $F_1$:  This metric is defined upon a token-level partial match score Intersection-Over-Union (IOU). For two spans $a$ and $b$, IOU is the quotient of the number of their overlapped tokens  and the number of their union tokens: $\text{IOU}=\frac{|a\cap b|}{|a\cup b|}$. If the IOU value between a rationale prediction and a ground truth rationale is above $0.5$, we consider this prediction as correct. Then, the $F_1$ score is calculated accordingly as the  IOU $F_1$.
    \item Token $P$, Token $R$, Token $F_1$:  For two spans, prediction rationale span $a$ and ground-truth rationale span $b$, token-level precision is the quotient of the number of their overlapped tokens  and the number of tokens in the prediction rationale span: $P_\text{token}=\frac{|a\cap b|}{|a|}$. The token-level recall is the quotient of the number of their overlapped tokens  and the number of tokens in the ground-truth rationale span: $R_\text{Token}=\frac{|a\cap b|}{|b|}$. Then, token $F_1$ is calculated as $\frac{2P_\text{token}R_\text{Token}}{P_\text{token}+R_\text{Token}}$.
    \item AUPRC: This metric is the area under the  precision ($P_\text{token}$)-recall ($R_\text{Token}$) curve. The calculate method is sweeping the threshold over the token-level scores.
    \item Comprehensiveness (Comp.): This metric means to judge whether the selected rationale is complete. To calculate this, we create a contrast example for each example  by removing the rationale $\bfz_\text{sym}$ from the original input $\bfx$, denoted by $\bfx/\bfz_\text{sym}$. After removing the rationales, the model should become less confident to the original predicted class $y$. We then measure comprehensiveness as follows: $\text{Comp.} = p(y|\bfx)-p(y|\bfx/\bfz_\text{sym})$. A high comprehensiveness score suggest that the extracted rationale is indeed complete for the prediction. 
    \item Sufficiency (Suff.): This metric means to judge whether the  selected rationale is useful. Similar to the comprehensiveness score,  we calculate the sufficiency score as: $\text{Suff.} = p(y|\bfx)-p(y|\bfz_\text{sym})$. If the extracted rationale is indeed useful, then the sufficiency score should be very small.
\end{itemize}

Among them, Token $P$, Token $R$, Token $F_1$,  IOU $F_1$, and AUPRC requires the gold rationale annotations, so we just calculate these metrics in the beer review task and the hate speech explaination task.  Comp. and Suff. only  fit for classification problems, so we just apply these metrics to legal judgment prediction task and hate speech explaination task.

\subsection{Beer Reviews}\label{sec:beer}

\subsubsection{Data.}
To provide a quantitative analysis for the extracted rationales, we use the BeerAdvocate\footnote{\url{https://www.beeradvocate.com/}} dataset~\cite{mcauley2012learning}. This dataset contains instances of human-written multi-aspect reviews on beers. Similarly to \citet{lei2016rationalizing}, we consider the following three aspects: appearance, smell, and palate. \newcite{mcauley2012learning} provide manually annotated rationales for 994 reviews for all aspects, which we use as test set.

The training set of BeerAdvocate contains 220,000 beer reviews, with human ratings for each aspect. 
Each rating is on a scale of $0$ to $5$ stars, and it can be fractional (e.g., 4.5 stars), \newcite{lei2016rationalizing} have normalized the scores to $[0,1]$, and picked ``less correlated'' examples to make a de-correlated subset.\footnote{\url{http://people.csail.mit.edu/taolei/beer/}} For each aspect, there are  80k--90k reviews for training and 10k reviews for validation.

\subsubsection{Model details.}
Because our task is a regression, we make some modifications to our model. First, we replace the \textit{softmax} in Eq.~\ref{eq:pyn} by the \textit{sigmoid} function, and replace the cross-entropy loss in Eq.~\ref{eq:lsp} by a mean-squared error (MSE) loss. Second, for a fair comparison, similar to \newcite{lei2016rationalizing} and \newcite{bastings2019interpretable}, we set all the architectures of selector, predictor, and guider as bidirectional Recurrent Convolution Neural Network (RCNN)~\newcite{lei2016rationalizing}, which performs similarly to an LSTM~\cite{hochreiter1997long} but with $50\%$ fewer parameters.

We search the hyperparameters in the following scopes: $\lambda_\text{ib}\in(0.000, 0.001]$ with step $0.0001$, $\lambda_g\in[0.2,2.0]$ with step $0.2$, $\lambda_\text{mi}\in[0.0, 1.0]$ with step $0.1$, and $\lambda_\text{lm}\in[0.000,0.010]$ with step $0.001$.

The best hyperparameters were found as follows: $\lambda_\text{ib}=0.0003$, $\lambda_g=1$, $\lambda_\text{mi}=0.1$, and $\lambda_\text{lm}=0.005$. 

We set $r_\phi(z_i)$  to $r_\phi(z_i=0)=0.999$ and $r_\phi(z_i=1)=0.001$.


\subsubsection{Evaluation Metrics and Baselines.}
For the evaluation of the selected tokens as rationales, we use precision, recall, and F1-score. Typically, precision is defined as the percentage of selected tokens that also belong to the human-annotated rationale. Recall is the percentage of human-annotated rationale tokens that are selected by our model. The predictions made by the selected rationale tokens are evaluated using the mean-square error (MSE).

We compare our method with the following baselines:
\begin{itemize}
\item Attention~\cite{lei2016rationalizing}: This method calculates attention scores over the tokens and selects top-k percent tokens as the rationale.
\item Bernoulli~\cite{lei2016rationalizing}: This method uses a selector network to calculate a Bernoulli distribution for each token, and then samples the tokens from the distributions as the rationale.  The basic architecture is RCNN~\newcite{lei2016rationalizing}.
\item HardKuma~\cite{bastings2019interpretable}:  This method replaces the Bernoulli distribution by a Kuma distribution to facilitate differentiability. The basic architecture is also  RCNN~\newcite{lei2016rationalizing}.
\item FRESH~\cite{jain-etal-2020-learning}: This method breaks the selector-predictor model into three sub-components: a support model which calculates the importance of each input token, a rationale extractor model which extracts the rationale snippets according to the output of the support model, a classifier model which make prediction according to the extracted rationale.
\item Sparse IB~\cite{paranjape-etal-2020-information}: This method also uses information bottleneck to control the number of tokens selected by the rationale. But it did not use any information calibration methods or any regularizers  to extract more complete and fluent rationales. 
\end{itemize}



\subsubsection{Results.}
The rationale extraction performances are shown in Table~\ref{tab:beer_rational}. The precision values for the baselines are directly taken from \cite{bastings2019interpretable}. We use their source code for the Bernoulli\footnote{\url{https://github.com/taolei87/rcnn}} and HardKuma\footnote{\url{https://github.com/bastings/interpretable_predictions}} baselines. 
%

\begin{figure}[!t]
\centering
\includegraphics[width=0.75\linewidth]{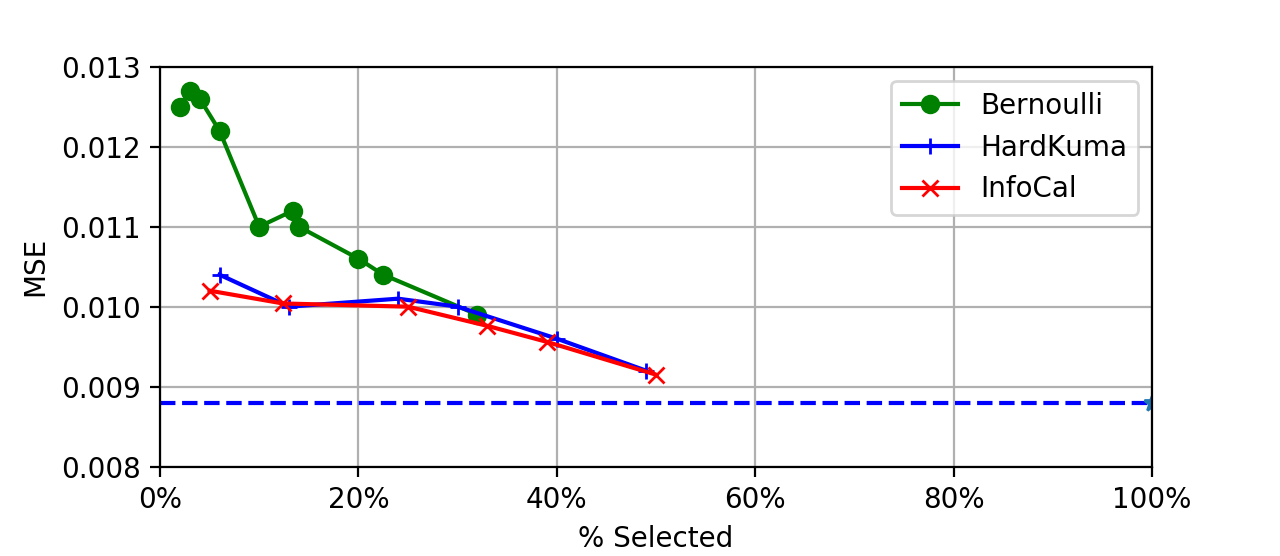}\\

\caption{MSE of all aspects of BeerAdvocate. The blue dashed line represents the full-text baseline (all tokens are selected).}
\label{tab:mse}
\end{figure}
\begin{table}[!t]
\centering
\resizebox{0.49\linewidth}{!}{
\begin{tabular}{|l|c|c|c|c|c||c|}
\hline
\multirow{2}{*}{Method} & \multicolumn{6}{c|}{Appearance} \\\cline{2-7}
           & P& R& F &IOU $F_1$& \% selected  & AUPRC    \\\hline
   Attention  &80.6   &35.6   & 49.4  &32.8 &13&0.613 \\\hline
   Bernoulli  &96.3   &56.5   &71.2     &55.3&14&0.785 \\\hline
   HardKuma  &98.1   &65.1    &78.3     &64.3&13&0.833 \\\hline
   FRESH  &96.5   &53.2    &  68.6     &52.2&13&0.772 \\\hline
   Sparse IB  & 91.3  &  54.6  & 68.3    & 51.9    &13&0.752   \\\hline\hline
   InfoCal  &\textbf{98.5}   &\textbf{73.2} &\textbf{84.0}  &72.4&13& 0.871 \\\hline
\end{tabular}
}
\resizebox{0.49\linewidth}{!}{
\begin{tabular}{|l|c|c|c|c|c||c|}
\hline
\multirow{2}{*}{Method}   & \multicolumn{6}{c|}{Smell} \\\cline{2-7}
           & P& R& F &IOU $F_1$& \% selected  & AUPRC    \\\hline
   Attention    &88.4  &20.6 &33.4    &20.1&7&   0.584\\\hline
   Bernoulli    &95.1  &38.2 &54.5   &37.5&7 & 0.697\\\hline
   HardKuma     &\textbf{96.8}  &31.5 &47.5    &31.2&7& 0.675 \\\hline
   FRESH    &90.4  &32.3 &47.6     &31.2  &7 &0.647 \\\hline
   Sparse IB  & 90.8  &  34.5  &  50.0   &  33.3   &7& 0.659  \\\hline\hline
   InfoCal    &95.6  &\textbf{45.6} &\textbf{61.7}      &44.7&7& 0.733\\\hline
\end{tabular}
}
\vspace{5pt}

\resizebox{0.49\linewidth}{!}{
\begin{tabular}{|l|c|c|c|c|c||c| }
\hline
\multirow{2}{*}{Method} &   \multicolumn{6}{c|}{Palate}\\\cline{2-7}
          & P& R& F&IOU $F_1$& \% selected   & AUPRC \\\hline
  Attention  &65.3  &35.8  &46.2   &30.1&7&0.537  \\\hline
  Bernoulli     &80.2  &53.6  &64.3 &47.3&7& 0.692\\\hline
  HardKuma      &\textbf{89.8}  &48.6  &63.1  &46.1&7&0.718 \\\hline
  FRESH    &78.4  &50.2  &61.2   &44.1&7
  & 0.668\\\hline
  Sparse IB  & 84.3  &  49.2 & 62.1   &  45.1   &7&  0.692 \\\hline\hline
  InfoCal     &89.6  &\textbf{59.8}  &\textbf{71.7}  &55.9&7& 0.767\\\hline
\end{tabular}
}
\smallskip 
 \caption{Token-level precision (P), recall (R),  F1-score (F),  IOU $F_1$, and AUPRC of selected rationales for the three aspects of BeerAdvocate. In bold, the best performance. ``\% selected'' means the average percentage of tokens selected out of the total number of tokens per instance.}
\label{tab:beer_rational}
\end{table}%

\begin{table}[!t]
\centering
\begin{tabular}{|l|c|c|c|c|c|c|c|c|c|}
\hline
\multirow{2}{*}{Method} & \multicolumn{3}{c|}{Appearance} & \multicolumn{3}{c|}{Smell}& \multicolumn{3}{c|}{Palate}\\\cline{2-10}
           & P& R& F    &  P& R& F       &  P& R& F  \\\hline
   InfoCal (HardKuma reg)  &97.9   &71.7 &  82.8    &94.8  &42.3 & 58.5        &89.4  &56.9  & 69.5     \\\hline
   InfoCal (INVASE reg)  &  96.8  &53.5 &  68.9       &93.2  &35.7 &51.6         &85.7  &39.5 & 54.1    \\\hline
   InfoCal$-L_\text{adv}$  &97.3      &67.8   & 79.9  &94.3  &34.5 &50.5    &89.6  &51.2  &65.2 \\\hline
   InfoCal$-L_\text{lm}$  &79.8   &54.9    &65.0   &87.1  &32.3 &47.1    &83.1  &47.4  &60.4  \\\hline
   InfoCal  &\textbf{98.5}   &\textbf{73.2} &\textbf{84.0}   &95.6  &\textbf{45.6} &\textbf{61.7}    &89.6  &\textbf{59.8}  &\textbf{71.7} \\\hline
\end{tabular}
\smallskip 
 \caption{The ablation tests. All the listed methods are tuned to select $13\%$ words in ``Appearence'', $7\%$ in ``Smell'' and ``Palate'' to make them comparable with InfoCal. The best performances are bolded.}
\label{tab:beer_rational_ablation}
\end{table}%

\begin{table}[!t]
\centering
\resizebox{\linewidth}{!}{
\begin{tabular}{|c|m{15cm}|}
\hline
Gold & \textcolor{red}{clear , burnished copper-brown topped by a large beige head that displays impressive persistance and leaves a small to moderate amount of lace in sheets when it eventually departs}
\textcolor{green}{the nose is sweet and spicy and the flavor is malty sweet , accented nicely by honey and by abundant caramel/toffee notes .} there ...... alcohol . 
\textcolor{blue}{the mouthfeel is exemplary ; full and rich , very creamy . mouthfilling with some mouthcoating as well .} drinkability is high ......
\\\hline
Bernoulli & \textcolor{red}{clear , burnished copper-brown topped by a large beige head that displays impressive persistance and} leaves a small to moderate amount of lace in sheets when it eventually departs
the nose is \textcolor{green}{sweet and spicy} and the flavor is malty sweet , accented nicely by honey and by abundant caramel/toffee notes . there ...... alcohol . 
the mouthfeel \textcolor{blue}{is exemplary ; full and rich , very creamy . mouthfilling with} some mouthcoating as well . drinkability is high ......\\\hline
HardKuma & \textcolor{red}{clear , burnished copper-brown topped by a large beige head that displays impressive persistance and leaves a small} to moderate amount of lace in sheets when it eventually departs the nose is \textcolor{green}{sweet and spicy} and the flavor is malty sweet , accented nicely by honey and by abundant caramel/toffee notes . there ...... alcohol . the mouthfeel \textcolor{blue}{is exemplary ; full and rich , very creamy . mouthfilling with} some mouthcoating as well . drinkability is high ...... \\\hline
InfoCal & \textcolor{red}{clear , burnished copper-brown topped by a large beige head that displays impressive persistance} and leaves a small to moderate amount of lace in sheets when it eventually departs \textcolor{green}{the nose is sweet and spicy and the flavor is malty sweet} , accented nicely by honey and by abundant caramel/toffee notes . there ...... alcohol . \textcolor{blue}{the mouthfeel is exemplary ; full and rich , very creamy . mouthfilling with some mouthcoating }as well . drinkability is high ...... \\\hline
InfoCal$-L_\text{adv}$ &\textcolor{red}{clear , burnished copper-brown topped by a large beige head that displays impressive persistance} and leaves a small to moderate amount of lace in sheets when it eventually departs the nose is \textcolor{green}{sweet and spicy} and the flavor is malty sweet , accented nicely by honey and by abundant caramel/toffee notes . there ...... alcohol . \textcolor{blue}{the mouthfeel is exemplary }; full and rich , very creamy . mouthfilling with some mouthcoating as well . drinkability is high ...... \\\hline
InfoCal$-L_\text{lm}$ & \textcolor{red}{clear , burnished} copper-brown topped by a large beige head that displays \textcolor{red}{impressive persistance} and leaves a small to \textcolor{red}{moderate amount of lace} in sheets when it eventually departs the nose is \textcolor{green}{sweet} and \textcolor{green}{spicy} and the flavor is \textcolor{green}{malty sweet} , accented nicely by \textcolor{green}{honey} and by abundant caramel/toffee notes . there ...... alcohol . the mouthfeel is \textcolor{blue}{exemplary} ; \textcolor{blue}{full} and rich , very \textcolor{blue}{creamy} . mouthfilling with some mouthcoating as well . drinkability is high ...... \\
\hline
\end{tabular}
}

\smallskip 
\caption{One example of extracted rationales by different methods. Different colors correspond to different aspects: \textcolor{red}{red}: appearance, \textcolor{green}{green}: smell, and \textcolor{blue}{blue}: palate.}
\label{tab:case}
\end{table}%

\begin{figure}[!t]
\centering
\includegraphics[width=0.9\linewidth]{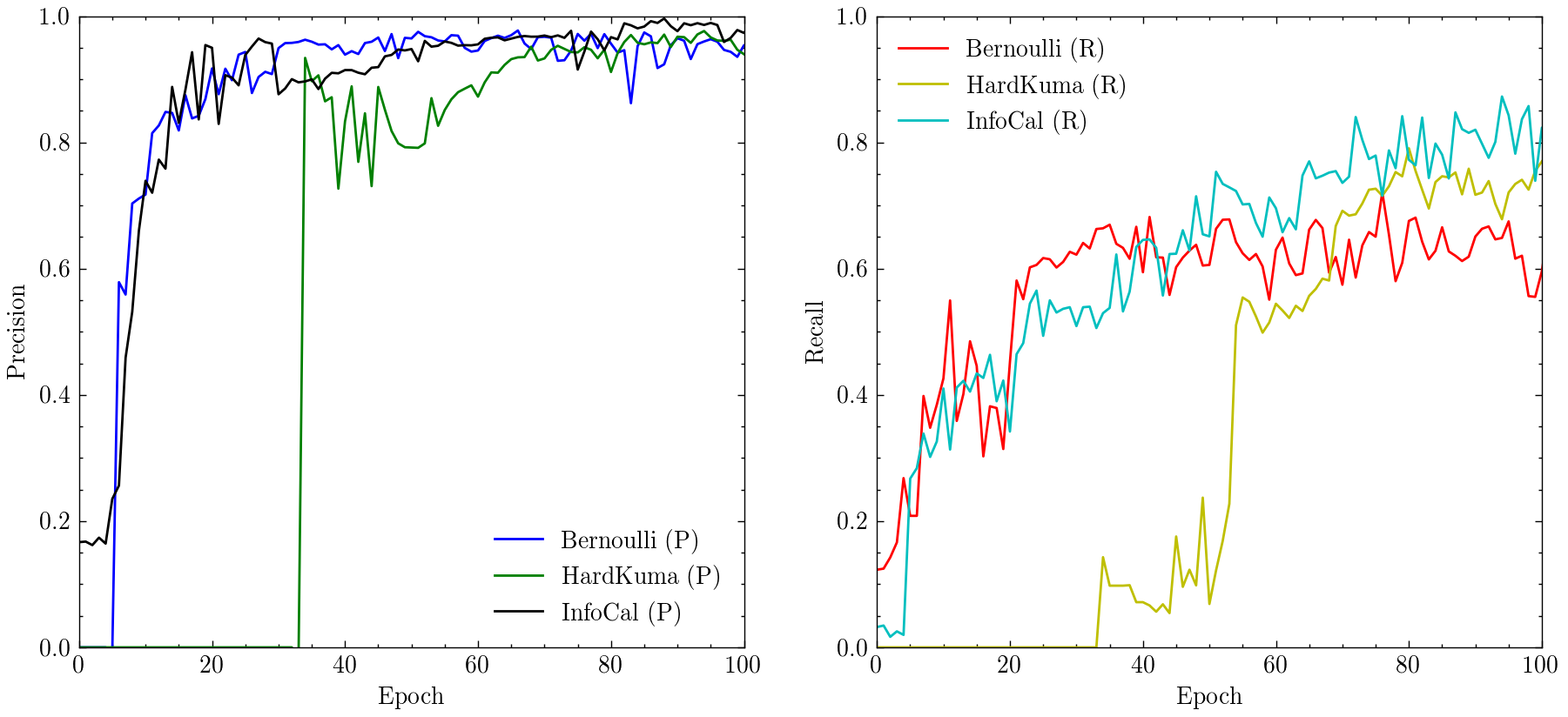}\\

\caption{The precision (left) and recall (right) for rationales  on the smell aspect of the BeerAdvocate valid set. } 
\label{tab:pr}\vspace*{1ex}
\end{figure}
We trained these baseline for 50 epochs and selected the models with the best recall on the dev set when the precision was equal or larger than the reported dev precision. For fair comparison, we used the same stopping criteria for InfoCal (for which we fixed a threshold for the precision at 2\% lower than the previous state-of-the-art). 

We also conducted ablation studies: (1) we removed the adversarial loss and report the results in the line  InfoCal$-L_\text{adv}$, and (2)~we removed the LM regularizer and report the results in the line InfoCal$-L_\text{lm}$. 
 
In Table~\ref{tab:beer_rational}, we see that, although Bernoulli, HardKuma,  FRESH, and Sparse IB achieve very high precisions, their recall scores are significantly low. 
The reason is that these four methods only focus on making the extracted rationale enough for a correct prediction, so the rationale is not necessary to be competent and many details would  lost. In comparison, our InfoCal method use a dense neural network as a guider, which  provided many detailed information. Therefore, the selector is able to extract more complete rationales.

In comparison, our method InfoCal significantly outperforms the previous methods in the recall scores for all the three aspects of the BeerAdvocate dataset (we performed Student's t-test, $p<0.01$). Also, all the three F-scores of InfoCal are a new state-of-the-art performance.

In the ablation studies in Table~\ref{tab:beer_rational_ablation}, we see that when we remove the adversarial information calibrating structure, namely, for InfoCal$-L_\text{adv}$, the recall scores decrease significantly in all the three aspects. This shows that our guider model is critical for the increased performance. 
Moreover, when we remove the LM regularizer, we find a significant drop in both precision and recall, in the line InfoCal$-L_\text{lm}$. This highlights the importance of semantical fluency of rationales, which are encouraged by our LM regularizer.  

We also apply another kind of calibration, which was applied in \newcite{yoon2018invase}. This calibration method is very similar to the ``base'' model in actor-critic models~\cite{konda2000actor}. Their difference with our InfoCal is that \newcite{yoon2018invase} minimizes the difference between the cross entropy values of  the selector-predictor model and the base model. We apply their method to our model and listed the results in the InfoCal (INVASE reg) line in Table~\ref{tab:beer_rational_ablation}. We found that the recall score decreases a lot compared to InfoCal, which shows that our information calibration method is better for improving the recall of rationale extraction.

We also replace the LM regularizer with the regularizer used in the  HardKuma method with all the other parts of the model unchanged, denoted InfoCal(HardKuma reg) in Table~\ref{tab:beer_rational_ablation}. We found that the recall and F-score of InfoCal outperforms InfoCal(HardKuma reg), which shows the effectiveness of our LM regularizer.

\begin{table}[!t]
\begin{center}
\resizebox{\linewidth}{!}{
\begin{tabular}{l l |ccccc| ccccc |ccccc}
\toprule[1.0pt]
\multirow{2}{*}{Small}&Tasks &  \multicolumn{5}{c|}{Law Articles} & \multicolumn{5}{c|}{Charges} & \multicolumn{5}{c}{Terms of Penalty}\\
\cmidrule[0.5pt]{2-17}
 &Metrics& Acc & MP & MR & F1 & \%S & Acc & MP & MR & F1 &\%S & Acc & MP & MR & F1 &\%S   \\
\midrule[0.5pt]
\multirow{8}{*}{Single} &Bernoulli (w/o) &0.812 & 0.726 & 0.765 & 0.756 &100 & 0.810 & 0.788 & 0.760 & 0.777  & 100 &  0.331 & 0.323 & 0.297 & 0.306 & 100\\
&Bernoulli &0.755 & 0.701 & 0.737 & 0.728 &14 & 0.761 & 0.753 & 0.739 & 0.754  & 14 &  0.323 & 0.308 & 0.265 & 0.278 & 30\\
  &HardKuma (w/o) & 0.807 & 0.704 & 0.757 & 0.739 &100&0.811 & 0.776 & 0.763 & 0.776   &100 &  0.345 & 0.355 & 0.307 & 0.319& 100\\
  &HardKuma & 0.783 & 0.706 & 0.735 & 0.729 &14&0.778 & 0.757 &0.714 &0.736 &14 &  0.340 & 0.328 &0.296 & 0.309 & 30\\
  &FRESH & 0.801 & 0.714 & 0.761 & 0.743 &14&0.790 & 0.766 &0.725 &0.745 &14 &  0.344 & 0.332 &0.308 & 0.312 & 30\\
  &Sparse IB &0.773 & 0.692 & 0.734 &  0.712   &14 & 0.769 & 0.758 & 0.742 & 0.750     & 14 &  0.336 & 0.324 & 0.280   & 0.300   & 30\\
 \cline{2-17}
 &InfoCal$-L_\text{adv}$ & 0.826  &0.739 &0.774 &0.777 &14 &0.845  &0.804 &0.781 &0.797  &14  &0.351 &0.374 & 0.329&0.330 &30\\
 &InfoCal$-L_\text{adv}\!\!-\!\!L_\text{ib}$ (w/o) & \textbf{0.841}  & \textbf{0.759}& \textbf{0.785}&\textbf{0.793} &100  &  \textbf{0.850}  & \textbf{0.820}& \textbf{0.801}&\textbf{0.814}  &100 & \textbf{0.368}&\textbf{0.378} &\textbf{0.341}  & \textbf{0.346}&100\\
 &InfoCal$-L_\text{lm}$ & 0.822  &0.723 &0.768 &0.773 &14 &0.843  &0.796 &0.770 &0.772 &14  &0.347 &0.361 & 0.318&0.320 &30\\
 &InfoCal              & \uwave{0.834}  &\uwave{0.744} &\uwave{0.776} &\uwave{0.786} &14 &\uwave{0.849}  &\uwave{0.817} &\uwave{0.798} &\uwave{0.813} &14  &\uwave{0.358} &\uwave{0.372} & \uwave{0.335}&\uwave{0.337} &30\\
 \midrule[0.5pt]
\multirow{3}{*}{Multi} &FLA & 0.803& 0.724& 0.720 &0.714 &$-$&0.767& 0.758& 0.738& 0.732&$-$ &0.371& 0.310 &0.300 &0.299&$-$\\
 &TOPJUDGE & 0.872 & 0.819 &0.808 &0.800 &$-$&0.871 &0.864& 0.851 &0.846&$-$& 0.380 &0.350 &0.353&0.346&$-$\\
 &MPBFN-WCA &\underline{0.883} &\underline{0.832}& \underline{0.824} &\underline{0.822}&$-$ &\underline{0.887} &\underline{0.875} &\underline{0.857} &\underline{0.859}&$-$&\underline{ 0.414} &\underline{0.406} &\underline{0.369}& \underline{0.392}&$-$\\

\midrule[1.0pt]
\multirow{2}{*}{Big}&Tasks &  \multicolumn{5}{c|}{Law Articles} & \multicolumn{5}{c|}{Charges} & \multicolumn{5}{c}{Terms of Penalty}\\
\cmidrule[0.5pt]{2-17}
 &Metrics& Acc & MP & MR & F1 &\%S& Acc & MP & MR & F1&\%S & Acc & MP & MR & F1 &\%S\\
\midrule[0.5pt]
 \multirow{8}{*}{Single} &Bernoulli (w/o) & 0.876    &0.636   & 0.388 &0.625 &100 & 0.857      &0.643    &0.410   &0.569 &100 & 0.509    &0.511   &0.304  &0.312 & 100\\
 &Bernoulli & 0.857    &0.632   & 0.374 &0.621 &14 & 0.848      &0.635    &0.402   &0.543 &14 & 0.496    &0.505   &0.289  &0.306 & 30\\
  &HardKuma (w/o)&  0.907 & 0.664 & 0.397 & 0.627 & 100&  0.907 & 0.689 & 0.438 & 0.608&100 &   0.555 & 0.547 & 0.335 & 0.356&100\\
  &HardKuma &  0.876 & 0.645 & 0.384 & 0.609 & 14&  0.892 & 0.676 & 0.425 & 0.587&14 &   0.534 & 0.535 & 0.310 & 0.334&30\\
  &FRESH &  0.902 & 0.698 & 0.675 & 0.682 & 14&  0.902 & 0.695 & 0.632 & 0.653&14 &   0.532 & 0.539 & 0.343 & 0.387&30\\
  &Sparse IB & 0.863     &0.634   & 0.372 & 0.624   &14 & 0.852      &0.638    &0.401   & 0.545     &14 & 0.501    &0.510   &0.286  &  0.302    & 30\\
 \cline{2-17} 
 
 &InfoCal$-L_\text{adv}$ & 0.953 & 0.844& 0.711&0.782&20  &0.954& 0.857 & 0.772 &0.806& 20&0.552&0.490& 0.353& 0.356 &30\\
 &InfoCal$-L_\text{adv}\!\!-\!\!L_\text{ib}$ (w/o) &  \textbf{0.959} & \textbf{0.862} & \textbf{0.751}&0.791&100  &\textbf{0.957}&\textbf{0.878}&0.776&0.807&100 &\textbf{0.584}& \textbf{0.519}& \textbf{0.411}&\textbf{0.427} &30\\
 &InfoCal$-L_\text{lm}$ & 0.953 & 0.851& 0.730 & 0.775& 20 & 0.950& 0.857&0.756& 0.789 &20 &0.563& 0.486& 0.374& 0.367& 30\\
 &InfoCal & \uwave{0.956}&\uwave{0.852}& \uwave{0.742}& \textbf{0.805} & 20 &\uwave{0.955}&\uwave{0.868}& \textbf{0.788}&\textbf{0.820} &20 & 0.556&\textbf{0.519}&0.362&0.372 &30  \\
\midrule[0.5pt]
\multirow{3}{*}{Multi}&FLA & 0.942&0.763&0.695&0.746&$-$&0.931&0.798&0.747&0.780&$-$&0.531&0.437&0.331&0.370&$-$\\
&TOPJUDGE&0.963&0.870&0.778&0.802&$-$&0.960&0.906&0.824&0.853&$-$&0.569&0.480&0.398&0.426&$-$\\
&MPBFN-WCA&\underline{0.978}&\underline{0.872}&\underline{0.789}&\underline{0.820}&$-$&\underline{0.977}&\underline{0.914}&\underline{0.836}&\underline{0.867}&$-$&\underline{0.604}&\underline{0.534}&\underline{0.430}&\underline{0.464}&$-$\\
\bottomrule[1.0pt]
\end{tabular}
}
\end{center}
\caption{The overall performance on the CAIL2018 dataset (Small and Big). The results from previous works are directly quoted from \newcite{yang2019legal}, because we share the same experimental settings, and hence we can make direct comparisons. \%S represents the selection percentage (which is determined by the model). ``Single'' represents single-task models, ``Multi'' represents multi-task models. The best performance is in bold. The red numbers mean that they are less than the best performance by no more than $0.01$. The underlined numbers are the state-of-the-art performances, all of which are obtained by multi-task models. (w/o) represents that the corresponding model is a dense model without extracting rationales.}
\label{tab:overall_law}
\end{table}%

We further show the relation between a model's performance on predicting the final answer and the rationale selection percentage (which is determined by the model) in Fig.~\ref{tab:mse}, as well as the relation between precision/recall and training epochs in Fig.~\ref{tab:pr}. The rationale selection percentage is influenced by $\lambda_\text{ib}$. According to Fig.~\ref{tab:mse}, our method InfoCal achieves a similar prediction performance compared to previous works, and does slightly better than HardKuma for some selection percentages. Fig.~\ref{tab:pr} shows the changes in precision and recall with training epochs. We can see that our model achieves a similar precision after several training epochs, while significantly outperforming the previous methods in recall, which proves the effectiveness of our proposed method.
 
Table~\ref{tab:case} shows an example of rationale extraction. Compared to the rationales extracted by Bernoulli and HardKuma, our method provides more fluent rationales for each aspect. For example, unimportant tokens like ``and'' (after ``persistance'', in the Bernoulli method), and ``with'' (after ``mouthful'', in the HardKuma method) were selected just because they are adjacent to important ones.

\subsection{Legal Judgement Prediction}\label{sec:law}
\subsubsection{Datasets and Preprocessing.}
We use the CAIL2018 data\-set\footnote{ \url{https://cail.oss-cn-qingdao.aliyuncs.com/CAIL2018_ALL_DATA.zip}}~\cite{zhong-etal-2018-legal} for three tasks on legal judgment prediction.
The dataset consists of criminal cases published by the Supreme People's Court of China.\footnote{\url{http://cail.cipsc.org.cn/index.html}} To be consistent with previous works, we used two versions of CAIL2018, namely, CAIL-small (the exercise stage data) and CAIL-big (the first stage data). The statistics of CAIL2018 dataset are shown in Table~\ref{tab:cail}.

The instances in CAIL2018 consist of a \textit{fact description} and three kinds of annotations: \textit{applicable law articles}, \textit{charges}, and \textit{the penalty terms}. Therefore, our three tasks on this dataset consist of predicting (1)~law articles, (2)~charges, and (3)~terms of penalty according to the given fact description.

\begin{table}[!t]
\begin{center}
\begin{tabular}{lcc}
\toprule[1.0pt]
 & CAIL-small & CAIL-big\\
\midrule[0.5pt]
Cases & 113,536 & 1,594,291\\
Law Articles & 105 &183 \\
Charges & 122 & 202\\
Term of Penalty &11 & 11\\
\bottomrule[1.0pt]
\end{tabular}
\end{center}
\caption{Statistics of the CAIL2018 dataset.}
\label{tab:cail}
\end{table}%

In the dataset, there are also many cases with multiple applicable law articles and multiple charges. To be consistent with previous works on legal judgement prediction \cite{zhong-etal-2018-legal,yang2019legal}, we filter out these multi-label examples. 

We also filter out instances where the charges and law articles occurred less than $100$ times in the dataset (e.g., insulting the national flag and national emblem).
For the term of penalty, we divide the terms into $11$ non-overlapping intervals. These preprocessing steps are the same as in \newcite{zhong-etal-2018-legal} and \newcite{yang2019legal}, making it fair to compare our model with previous models.
 
We use Jieba\footnote{\url{https://github.com/fxsjy/jieba}} for token segmentation, because this dataset is in Chinese. The word embedding size is set to $100$ and is randomly initiated before training. The maximum sequence length is set to $1000$. The architectures of the selector, predictor, and guider are all bidirectional LSTMs. The LSTM's hidden size is set to $100$. $r_\phi(z_i)$ is the sampling rate for each token (0 for selected), which we set to $r_\phi(z_i=0)=0.9$. 

We search the hyperparameters in the following scopes:
$\lambda_{ib} \in [0.00, 0.10]$ with step $0.01$, $\lambda_g \in [0.2, 2.0]$
with step $0.2$, $\lambda_{mi} \in [0.0, 1.0]$ with step $0.1$, $\lambda_{lm} \in
[0.000, 0.010]$ with step $0.001$.
The best hyperparameters were found to be: $\lambda_{ib}=0.05, \lambda_{g}=1, \lambda_{mi}=0.5, \lambda_{lm}=0.005$ for all the three tasks. 

\subsubsection{Overall Performance.}
We again compare our method with the Bernoulli~\cite{lei2016rationalizing} and the  HardKuma~\cite{bastings2019interpretable} methods on rationale extraction. These two methods are both single-task models, which means that we train a model separately for each task. 
We also compare our method with three multi-task methods listed as follows:
\begin{itemize}
\item FLA~\cite{luo-etal-2017-learning} uses an attention mechanism to capture the interaction between fact descriptions and applicable law articles.
\item TOPJUDGE~\cite{zhong-etal-2018-legal} uses a topological architecture to link different legal prediction tasks together, including the prediction of law articles, charges, and terms of penalty.
\item MPBFN-WCA~\cite{yang2019legal} uses a backward verification to verify upstream tasks given the results of downstream tasks.
\end{itemize}
The results 
are listed in Table~\ref{tab:overall_law}.  

On CAIL-small, we observe that it is more difficult for the single-task models to outperform multi-task methods. This is likely due to the fact that the tasks are related, and learning them together can help a model to achieve better performance on each task separately. After removing the restriction of the information bottleneck,  InfoCal$-L_\text{adv}\!\!-\!\!L_\text{ib}$ achieves the best performance in all tasks, however,  it selects all the tokens in the review. 
When we restrict the number of selected tokens to $14\%$ (by tuning the hyperparameter $\lambda_\text{ib}$), InfoCal (in red) only slightly drops in all evaluation metrics, and it already outperforms Bernoulli and HardKuma, even if they have used all tokens. This  means that the $14\%$ selected tokens are very important to the predictions. We observe a similar phenomenon for CAIL-big. Specifically, InfoCal outperforms InfoCal$-L_\text{adv}\!\!-\!\!L_\text{ib}$  in some evaluation metrics, such as the F1-score of law article prediction and charge prediction tasks.

\subsubsection{Rationales.}

\begin{table}[]
    \centering
    \begin{tabular}{|c|c|c|c|c|c|c|}
    \hline
    &\multicolumn{2}{c|}{Law Articles} & \multicolumn{2}{c|}{Charges} & \multicolumn{2}{c|}{Terms of Penalty}\\
    \cline{2-7}
         & Comp.$\uparrow$ & Suff.$\downarrow$  &Comp.$\uparrow$  & Suff.$\downarrow$ &Comp.$\uparrow$  & Suff.$\downarrow$   \\
         \hline
       Bernoulli  & 0.231 &0.005 &0.243 & 0.002&0.132 &0.017\\
       HardKuma  & 0.304& -0.021&0.312 &-0.034 & 0.165&0.009\\
       InfoCal  &\textbf{0.395} & \textbf{-0.056}&\textbf{0.425} &\textbf{-0.067} &\textbf{0.203} &\textbf{0.005}\\
    \hline
    \end{tabular}
    \caption{The quantitative evaluation of rationales for legal judgment prediction. The ``$\uparrow$''
 means that a good result should have a larger value, while ``$\downarrow$'' means lower is better.}
    \label{tab:lawrationale}
\end{table}
The CAIL2018 dataset does not contain annotations of rationales. So, we only use Comp. and Suff. for quantitative evaluation since they do not require gold rationale annotations. The results are shown in Table~\ref{tab:lawrationale}. We can see that in all the three subtasks of legal judgement prediction, our proposed method outperforms the previous methods.

We also conducted human evaluation for the extracted rationales. Due to limited budget and resources, we sampled 300 examples for each task. We randomly shuffled the rationales for each task and asked six undergraduate students from Peking University to evaluate them. The human evaluation is based on three metrics: usefulness (U), completeness (C), and fluency (F); each scored from $1$ (lowest) to~$5$. The scoring standard for human annotators is given in Appendix \ref{human}  in the extended paper.  

The human evaluation results are  shown in Table~\ref{tab:he}. We can see that our proposed method outperforms previous methods in all metrics. Our inter-rater agreement is acceptable by Krippendorff's rule~(\citeyear{krippendorff2004content}), which is shown in Table~\ref{tab:he}.

\begin{figure}[!t]
\centering
\includegraphics[width=0.75\linewidth]{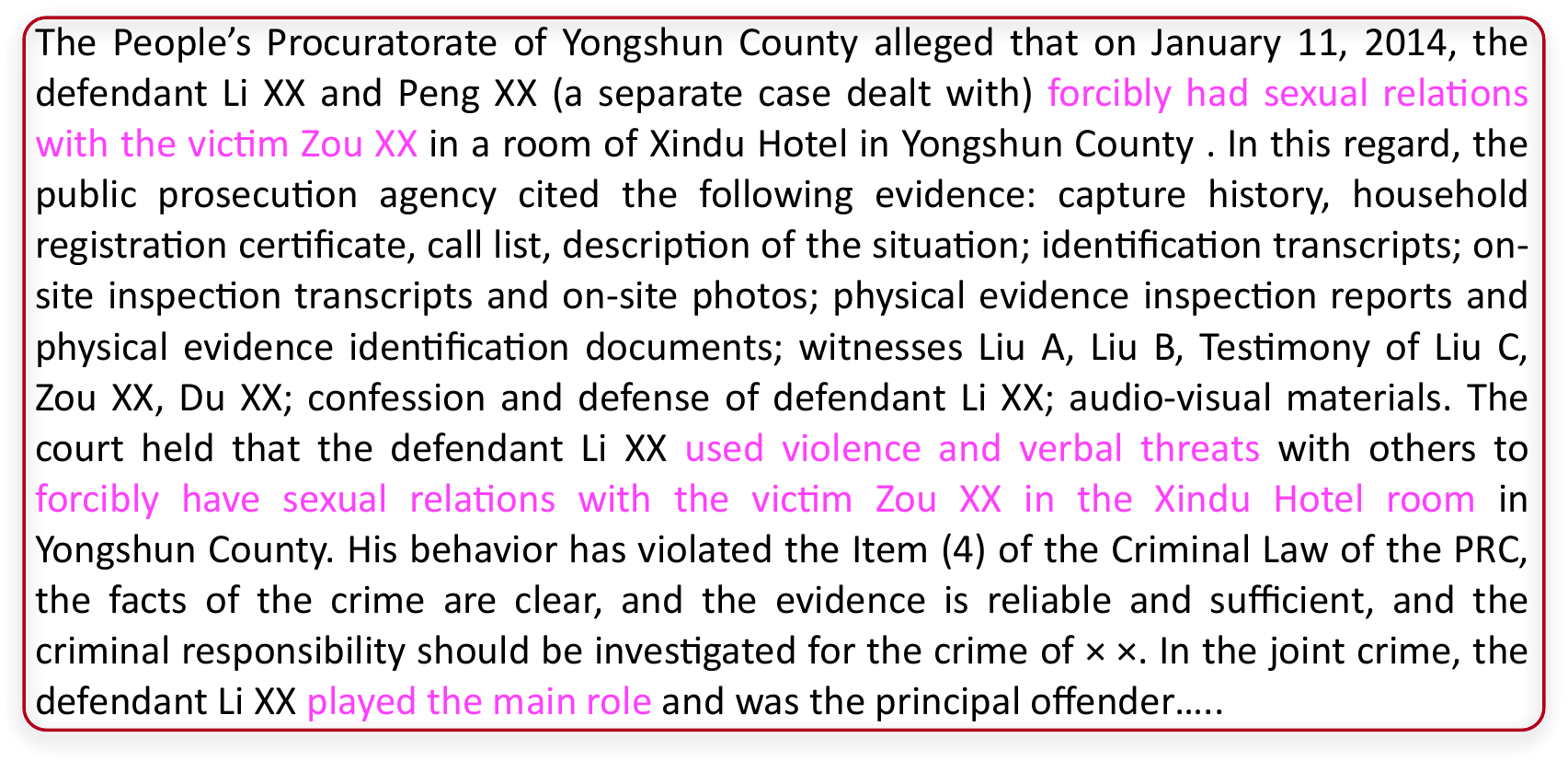}\\
\smallskip 
\caption{An example of extracted rationale for charge prediction. The correct charge is ``Rape". The original fact description is in Chinese, we have translated it to English. It is easy to see that the extracted rationales are very helpful in making the charge prediction.}
\label{fig:lawcase}
\end{figure}

A sample case of extracted rationales in legal judgement is shown in Fig.~\ref{fig:lawcase}. We observe that our method selects all the useful information for the charge prediction task, and the selected rationales are formed of continuous and fluent sub-phrases.

\begin{table}[!t]
\centering
\resizebox{0.75\linewidth}{!}{
\begin{tabular}{|l|c|c|c|c|c|c|c|c|c|}
\hline
         & \multicolumn{3}{|c|}{Law} &  \multicolumn{3}{|c|}{Charges} & \multicolumn{3}{|c|}{ToP} \\\cline{2-10}
          & U & C &F & U & C & F&U & C  & F \\\hline
Bernoulli &4.71  &2.46 &3.45 & 3.67 &2.35 &3.45 &3.35 &2.76&3.55 \\\hline
HardKuma  &4.65  &3.21 &3.78 & 4.01 &3.26&3.44  &3.84 &2.97&3.76\\\hline
InfoCal  &\textbf{4.72} &\textbf{3.78} &\textbf{4.02} &\textbf{4.65} & \textbf{3.89}&\textbf{4.23} &\textbf{4.21} &\textbf{3.43}&\textbf{3.97}\\\hline\hline
 $\alpha$&0.81 & 0.79&0.83&0.92&0.85&0.87&0.82&0.83&0.94\\\hline
\end{tabular}
}
\smallskip 
\caption{Human evaluation on the CAIL2018 dataset. ``ToP" is the abbreviation of ``Terms of Penalty". The metrics are: usefulness (U), completeness (C), and fluency (F), each scored from $1$ to~$5$. Best performance is in bold. $\alpha$ represents Krippendorff's alpha values. The basic architecture for the three methods are all RCNN~\newcite{lei2016rationalizing}.}
\label{tab:he}
\end{table}%

\subsection{Hate Speech Explanation}\label{sec:hate}
\subsubsection{Datasets and Preprocessing.}

For evaluating the performance of our method on hate speech detection task. We use the HateXplain dataset\footnote{\url{https://github.com/punyajoy/HateXplain.git}}~\cite{mathew2020hatexplain}. This dataset contains   9,055 posts from Twitter~\cite{davidson2017automated,fortuna2018survey} and  11,093 posts from Gab~\cite{lima2018inside,mathew2020hate,zannettou2018gab}.
There are three different classes in this dataset: hateful, offensive, and normal. Apart from the class labels, this dataset also contains rationale annotations for each example that is labelled as hateful or offensive. The training set, valid set, and test set are already split as $8:1:1$ in the dataset. 
More details of this dataset is shown in Table~\ref{tab:hatedata}. This dataset is very noisy,  and it can test the robustness of our InfoCal method on noisy text information.

For classification performance, we have three metrics: Accuracy, Macro $F_1$, and AUROC. These metrics are used for evaluating the ability of distinguish among the three classes, i.e., hate speech, offensive speech, and normal. Among them, AUROC is the area under the ROC curve. 

\subsubsection{Competing Methods.}
We also compare our method with Bernoulli~\cite{lei2016rationalizing} and HardKuma~\cite{bastings2019interpretable} in this experiment. 
We also compare our method with the following competing methods provided in \newcite{mathew2020hatexplain}:
\begin{itemize}
    \item \textbf{CNN-GRU}~\cite{zhang2018detecting} has achieved state-of-the-art performance in multiple hate speech datasets. CNN-GRU first use convolution neural network (CNN)~\cite{lecun1995convolutional} to capture the local features and then use recurrent neural network (RNN)~\cite{rumelhart1986learning} with GRU unit~\cite{cho2014learning} to capture the temporal information. Finally, this model max-pools GRU's hidden layers  to a feature vector, and then use a fully connected layer to finally output the prediction results. 
    \item \textbf{BiRNN}~\cite{schuster1997bidirectional} first input the tokens into a sequential model with long-short term memory (LSTM)~\cite{hochreiter1997long}. Then, the last hidden state is passed through two feed-forward layers and then a fully connected layer for prediction.
    \item \textbf{BiRNN-Attn} adds an attention layer after the sequential layer of BiRNN model.
    \item \textbf{BERT}~\cite{devlin-etal-2019-bert} is a large pretrained model constructed by a stack of transformer~\cite{vaswani2017attention} encoder layers. A fully connected layer is added to the output corresponding to the \textit{CLS} token for the hate speech class prediction. We used the  \texttt{bert-base-uncased} model with  12-layer, 768-
hidden, 12-heads, 110M parameters, this is the same setting with previous work~\cite{mathew2020hate}. The model is fine-tuned using the HateXplain training set.
\end{itemize}
In all the  above methods, the rationales are extracted by two methods: attention~\cite{rocktaschel2015reasoning} and LIME~\cite{ribeiro2016should}. When we are using attention method, as is described in \newcite{deyoung-etal-2020-eraser}, the tokens with  top 5 attention values are selected as rationale. The LIME method selects rationales by training a new  explanation model to imitate the original deep learning ``black-box'' model. Different from these methods, our model InfoCal as well as the other two competing method Bernoulli and HardKuma are extracting rationales by the model itself without  any external methods (like attention selection or LIME selection) for rationale selection. So, it is much more challenging for them to achieve similar explanability performance.

In \newcite{mathew2020hatexplain}, the ground-truth rationale annotations were also used to train some models by adding an external cross entropy loss on the attention layer. The rationale training is conducted on BiRNN and BERT models, denoted as BiRNN-HateXplain and BERT-HateXplain, respectively.

\subsubsection{Results.}
The overall results are shown in Table~\ref{tab:hate}. We can see that in the classification performance, the BERT models achieved the highest score in all the three metrics (Accuracy, Macro $F_1$, and AUROC) no matter whether the rationale supervising  is conducted. Also, our InfoCal model has outperformed all the other approaches except for BERT. This makes sense because BERT has pretrained by a large amount of texts, and it has a much better understanding for language than other models without pretraining. 

In the explanability evaluations, our model InfoCal has achieved the state-of-the-art performance in three metrics: IOU $F_1$, AUPRC, and Sufficiency. Also, for the other two metrics (Token $F_1$ and Comprehensiveness), the InfoCal method is comparable with the state-of-the-art method (BERT [Attn]). Note that in our model, the rationales are selected by the model itself instead of by  selecting top 5 attention value or by LIME method externally. Therefore, this experimental result show that our InfoCal model is a better model for explaining neural network predictions.

We also listed the performances of the BiRNN model and BERT model after supervised by rationale annotations in Table~\ref{tab:hate}. We can see that  both the classification performance and the explanability performance improved a lot after trained by  rationale annotations. This also makes sense because the rationale annotation is the most direct training signal of rationale selection. However, such kind of rationale annotation is very expensive to get in real-world applications. Therefore, the rationale extraction methods without rationale supervision is much proper to be applied in the industry.

\begin{table}[!ht]
    \centering
    \begin{tabular}{cccc}
    \toprule[1.0pt]
         &Twitter & Gab & Total  \\
    \midrule[0.5pt]
      Hateful   & 708 & 5,227 & 5,935\\
      Offensive &2,328 &3,152 &5,480\\
      Normal & 5,770& 2,044& 7,814\\
      Undecided &249& 670& 919\\
    \midrule[0.5pt]
      Total &9,055& 11,093 &20,148\\
    \bottomrule[1.0pt]
    \end{tabular}
    \caption{The statistics of HateXplain dataset. ``Undecided'' means that in the annotation process, all the three annotators gave different labels to the example. We omit this part of data in our experiments as is consistent with previous works. }
    \label{tab:hatedata}
\end{table}
 
\begin{table}[!t]
\centering
\resizebox{\linewidth}{!}{
\begin{tabular}{ll|ccc|ccccc}
\toprule[1.0pt]
&& \multicolumn{3}{c|}{Classification Performance}&\multicolumn{5}{c}{Explanability}\\
\midrule[0.5 pt]
&&Acc $\uparrow$ & Macro F1 $\uparrow$ & AUROC $\uparrow$ & IOU F1 $\uparrow$ & Token F1 $\uparrow$& AUPRC $\uparrow$ & Comp $\uparrow$& Suff $\downarrow$\\
\midrule[0.5 pt]
\multirow{9}{1.2cm}{W/o rationale supervising}&CNN-GRU [LIME] & 0.627& 0.606& 0.793& 0.167&0.385& 0.648&0.316& -0.082 \\
&BiRNN [LIME] & 0.595& 0.575 &0.767&0.162 &0.361 &0.605&0.421& -0.051 \\
&BiRNN-Attn [Attn] & 0.621& 0.614 &0.795&0.167 &0.369& 0.643&0.278& 0.001 \\
&BiRNN-Attn [LIME] & 0.621& 0.614 &0.795&0.162 &0.386& 0.650&0.308& -0.075 \\
&BERT [Attn] & \textbf{0.690} &\textbf{0.674} &\textbf{0.843}&0.130 &\textbf{0.497}& 0.778&\textbf{0.447} &0.057 \\
&BERT [LIME] & \textbf{0.690} &\textbf{0.674} &\textbf{0.843}&0.118 &0.468& 0.747&0.436 &0.008 \\
\cmidrule[0.5 pt]{2-10}
&Bernoulli &0.597 &0.568 &0.765 &0.138 &0.482&0.668&0.324&0.003 \\
&HardKuma & 0.594 &0.570 &0.772 &0.152 &0.485&0.672&0.406&-0.022 \\
&Sparse IB &0.602 &0.572 &0.768 &0.145 &0.486&0.670&0.389&0.001 \\
&InfoCal &0.630 &0.614 &0.792 &\textbf{0.206} &0.493&\textbf{0.680} &0.436&\textbf{-0.097 }\\
\midrule[0.5 pt]
\multirow{4}{1.2cm}{With rationale supervising}&BiRNN-HateXplain [Attn] & 0.629  &0.629& 0.805&\underline{0.222} &\underline{0.506}& \underline{0.841}&0.281& 0.039 \\
&BiRNN-HateXplain [LIME] &0.629 &0.629 &0.805&0.174 &0.407& 0.685&0.343& -0.075 \\
&BERT-HateXplain [Attn] &\underline{0.698}  &\underline{0.687}& \underline{0.851} &0.120 &0.411& 0.626&0.424& 0.160\\
&BERT-HateXplain [LIME] &\underline{0.698} &\underline{0.687}& \underline{0.851} &0.112 &0.452 &0.722&\underline{0.500} &0.004 \\
\bottomrule[1.0pt]
\end{tabular}
}
\smallskip 
\caption{The overall performance on the HateXplain dataset. The results from previous work are directly quoted from \newcite{mathew2020hatexplain}, because we share identical train/valid/test data split, and hence we can make direct comparison. The ``$\uparrow$'' means that a good result should have a larger value, while ``$\downarrow$'' means lower is better. The square bracket means the method of extracting rationales. Note that except for Bernoulli, HardKuma, Sparse IB, and InfoCal, all the other methods are extracting rationale by external methods ([Attn] or [LIME]). The basic architectures for Bernoulli, HardKuma, and InfoCal are all RCNN.}
\label{tab:hate}
\end{table}%

\subsubsection{Case Study for Rationales.}

In Table~\ref{tab:hatecase}, we have listed some of the generated rationales in HateXplain dataset by our InfoCal method and the two competing methods: Bernoulli and HardKuma. We can see that our InfoCal method has extracted nearly all of the annotated rationales in the ground-truth. Compared to Bernoulli and HardKuma, our InfoCal method do not extract nonsense rationales, such as ``yeah i also'' in the second line, and ``precinct and campaign meetings'' in the third line. This again shows the effectiveness of the information calibration method.

\begin{table}[]
    \centering
    \resizebox{\linewidth}{!}{
    \begin{tabular}{ p{4cm} p{4cm} p{4cm} p{4cm} }
    \toprule[1.0pt]
    Gold & Bernoulli & HardKuma & InfoCal\\
    \midrule[0.5pt]
    \textbf{jew faggot milo} is king of the alt right great it not the alt right anymore & \textbf{jew  faggot milo is} king of the alt right great it not the alt right anymore & \textbf{jew  faggot} milo is king of the alt right great it not the alt right anymore   &  \textbf{jew  faggot milo} is king of the alt right great it not the alt right anymore\\
     \midrule[0.5pt]
    yeah i also agree that the \textbf{middle eastern ones are coming mostly to hungary} but these other \textbf{refugees away from mudslimes} are all schengen regions why bother applying for refugee status when you can just move it perfectly legal & \textbf{yeah i also} agree that the middle eastern ones are coming mostly to hungary  but these other \textbf{refugees} away from mudslimes are all schengen regions why bother applying for refugee status when you can just move it perfectly legal &  yeah i also agree that the \textbf{middle eastern ones are coming mostly to hungary}  but these other \textbf{refugees} away from mudslimes \textbf{are all schengen regions} why bother applying for refugee status when you can just move it perfectly legal & yeah i also agree that the \textbf{middle eastern ones} are coming mostly to hungary  but these other \textbf{refugees away from mudslimes} are all schengen regions why bother applying for refugee status when you can just move it perfectly legal\\
     \midrule[0.5pt]
     precinct and campaign meetings  will confirm that it is only  whites who hold  any ideals about government \textbf{non whites want free govt cheese and jews are crazy for power}& precinct and campaign meetings will confirm that it is only \textbf{whites} who hold any ideals about government non whites want free govt cheese and \textbf{jews are crazy for power}& \textbf{precinct and campaign meetings}  will confirm that it is only  whites who hold  any ideals about government non whites want free  \textbf{govt cheese and jews are crazy for} power&  precinct and campaign meetings  will confirm that it is only \textbf{whites} who  hold  any ideals about government \textbf{non whites want free govt cheese and jews are crazy for power}\\
     \midrule[0.5pt]
     the labour party has  number  mps in prison for paedophilia  number  mps and councillors on the sex registry  number  million white kids \textbf{raped by paki muslims} all labour voters covered up by labour councils mps and a pm 
     & the labour party has \textbf{number} mps in prison for paedophilia \textbf{number} mps and councillors on the sex registry \textbf{number} million white kids raped \textbf{by paki muslims all} labour voters covered up by labour councils mps and a pm 
     & the labour party has number mps in prison for paedophilia number mps and councillors on the sex registry number million \textbf{white kids raped by paki muslims all labour voters} covered up by labour councils mps and a pm 
     & the labour party has number mps in prison for paedophilia number mps and councillors on the sex registry number million white \textbf{kids raped by paki muslims} all labour voters covered up by labour councils mps and a pm \\
     
    \bottomrule[1.0pt]
    \end{tabular}
    }
    \caption{Example rationales extracted by three methods: Bermoulli, HardKuma, and InfoCal. Note that in these cases,  many phrases are offensive or hateful. Nevertheless, this cannot be avoided due to the nature of the work.}
    \label{tab:hatecase}
\end{table}

\subsection{Performance of the Pretrained Language Model for the rationale regularizer}

\begin{table}[!ht]
    \centering
    \begin{tabular}{lccc}
    \toprule[1.0pt]
                   &  KenLM~\cite{heafield-2011-kenlm} & RNNLM~\cite{bengio2003neural,tomas2011rnnlm} & Our LM\\
    \midrule[0.5pt]
        Perplexity (Beer) &   66   & 50         &44\\
    \midrule[0.5pt]
        Perplexity (Legal Small) &   32   &  20        &
        29    \\
        Perplexity (Legal Big) & 11     & 69        &62
            \\
    \midrule[0.5pt]
        Perplexity (HateXplain) &  413    & 146     &165 \\
    \bottomrule[1.0pt]
    \end{tabular}
    \caption{The comparison of perplexity between language models.  }
    \label{tab:beer_lm}
\end{table}

In the InfoCal model, we need a pretrained language model (in Sec.~\ref{sec:LM}) for the rationale regularizer. Our language model described in Section~\ref{lmvec} is different from previous language model because it has to compute  probabilities for token's vector representations
instead of token's symbolic IDs. Therefore, the quality of the pretrained language model is paramount to the InfoCal model. In Table~\ref{tab:beer_lm}, we listed the comparison of the perplexity between our language model and two famous language models: Kenneth Heafield's language model (KenLM)~\cite{heafield-2011-kenlm} and recurrent neural network language model (RNNLM)~\cite{bengio2003neural,tomas2011rnnlm}. The training is conducted on the pure texts of the training data in the three tasks, and the trained models are tested on the pure texts of the corresponding test sets.
We can see that the perplexity of our language model is comparable to RNNLM and even better than kenLM in some datasets. This shows that the performance of our language model is acceptable to our experiments. We do not compare the perplexity with Transformer-based models like GPT~\cite{radford2018improving,radford2019language,brown2020language}, because these models usually use subword vocabularies (like Byte Pair Encoding (BPE)~\cite{radford2019language} and WordPiece~\cite{schuster2012japanese,devlin-etal-2019-bert} ) which makes the perplexities not comparable with our work. 

Also, from the comparison of perplexity score, we found that the perplexity of HateXplain dataset is obviously higher than the other two datasets, this shows that HateXplain dataset is very noisy. The results in Table~\ref{tab:hate} proves that our InfoCal model is able to extract sensitive rationales on noisy text data.

\section{Summary and Outlook}\label{sec:conclu}

In this work, we proposed a novel method to extract rationales for neural predictions. Our method uses an adversarial-based technique to make a selector-predictor model learn from a guider model. In addition, we proposed a novel regularizer based on language models, which makes the extracted rationales semantically fluent. In this way, the ``guider'' model  tells the selector-predictor model what kind of information (token) remains unselected or over-selected. 
We conducted experiments on a task of sentiment analysis, hate speech recognition and three tasks from the legal domain. According to the comparison between the extracted rationales and the gold rationale annotations in sentiment analysis task and hate speech recognition task, our InfoCal method improves the selection of rationales by a large margin. We also conducted ablation tests for the evaluation of the LM regularizer's contribution, which showed that our regularizer is effective in refining the rationales. 

As future work, the main architecture of our model can be directly applied to other domains, e.g., images or tabular data.  The image rationales can be applied in many read-world applications, such as medical image recognition~\cite{deruyver2009image} and automatic driving~\cite{reece1995control}.  Regularizers based on Manifold learning~\cite{cayton2005algorithms} is promising to be applied on image rationale extraction. The tabular rationales are very useful in some tasks like automatic disease diagnose~\cite{alkim2012fast}. When designing the regularizers for tabular rationales, a sensible method is to make use of  the relations between different fields of the tabular since  different kinds of data are closely related in medical experiment reports and many of them are potentially to contribute to the patients' diagnose result.
\section{Ethical Statement}

The paper does not present a new dataset. It also does not use demographic or identity characteristics information. Furthermore, the paper does not report on experiments that involve a lot of computing time/power. 
\begin{itemize}
 \item 	\textbf{Intended use.} While the paper presents an NLP legal prediction application, our method is not yet ready to be used in practice. Our work takes a step forward in the research direction of making legal prediction systems explainable, which should uncover the systems' potential biases and modes of failures, thus ultimately rendering them more reliable.  
 Thus, once it can be guaranteed a high likelihood of correctness and unbiasedness of the predictions and the faithfulness of their explanations w.r.t.\ the inner-working of the model, legal prediction systems may help to assist judges (and not replace them) in their decisions, so that they can process more cases, and more people can perceive justice than nowadays is the case. (At present, only a very small portion of cases is brought to court; especially poorer parts of the populations have essentially no access to the justice system, due to its high costs.) In addition, legal prediction systems may be used as second opinion and help to uncover mistakes or even biases of human judges. 
 Currently, legal prediction systems are being heavily researched in the literature without the explainability component that our paper is bringing. Hence, our approach is taking a step forward in assessing the reliability of the systems, although we do not currently guarantee the faithfulness of the provided explanations. Hence, our work is intended purely as a research advancement and not as a real-world tool. 
 
 
 \item 	\textbf{Failure modes.} 
 Our model may fail to provide correct and unbiased predictions and explanations that are faithfully describing its decision-making process. Ensuring correct and unbiased predictions as well as faithful explanations are very challenging open questions, and our work takes an important but far from final step forward in this direction.

 \item \textbf{Biases.} 
 If the training data contains biases, then a model may pick up on these biases, and hence it would not be safe to use it in practice. Our explanations may help to detect biases and potentially give insights to researchers on how to further develop models that avoid them. However, we do not currently guarantee the faithfulness of the explanations to the decision-making of the model.
 
 \item 	\textbf{Misuse potential.} As our method is not currently suitable for production, the legal prediction model should not be used in real-world legal judgement prediction tasks. 
 \item 	\textbf{Collecting data from users.} We do not collect data from users, we only use an existing dataset.
 \item 	\textbf{Potential harm to vulnerable populations.} Since our model learns from datasets, if there are under-represented groups in the datasets, then the model might not be able to learn correct predictions for these groups. However, our model provides explanations for its predictions, which may uncover the potential incorrect reasons for its predictions on under-represented groups. This could further unveil the under-representation of certain groups and incentivize the collection of more instances for such groups. However, we highlight again that our model is not yet ready to be used in practice and that it is currently a stepping stone in this important direction of research.

\end{itemize}

\section*{Acknowledgments}
 This work was supported by the ESRC grant ES/S010424/1 “Unlocking the Potential of AI for English Law”, an Early Career Leverhulme Fellowship, a JP Morgan PhD Fellowship, the Alan Turing Institute under the EPSRC grant EP/N510129/1, the AXA Research Fund, and the EU TAILOR grant 952215. We also acknowledge the use of Oxford’s Advanced Research Computing (ARC) facility, of the EPSRC-funded Tier 2 facility JADE (EP/P020275/1), and of GPU computing support by Scan Computers
International Ltd.

\printcredits
\bibliographystyle{cas-model2-names}
\bibliography{refs}
\clearpage

\appendix
\section*{Appendices}
\section{Proofs}
\subsection{Derivation of $I(\tilde{\bfz}_\text{sym},y)$}\label{the:0}
This proof is the basis for the information bottleneck equation in Section~\ref{sec:ib}. 
\begin{theorem}
Minimizing $-I(\tilde{\bfz}_\text{sym},y)$ is 
equivalent to minimizing $L_{sp}$.
\end{theorem}
\begin{proof}
\begin{equation}\label{eq:Izy}
I(\tilde{\bfz}_\text{sym},y)=\mathbb E_{\tilde{\bfz}_\text{sym},y}\Big[\frac{p(y|\tilde{\bfz}_\text{sym})}{p(y)}\Big]=\mathbb E_{\tilde{\bfz}_\text{sym},y}p(y|\tilde{\bfz}_\text{sym}) - \mathbb E_{\tilde{\bfz}_\text{sym},y}p(y).\\
\end{equation}
We omit $\mathbb E_{\tilde{\bfz}_\text{sym},y}p(y)$, because it is a constant, therefore, minimizing  Eq.~\ref{eq:Izy} is equivalent to minimizing the following term:
\begin{equation}\label{eq:Ezy}
\mathbb E_{\tilde{\bfz}_\text{sym},y}p(y|\tilde{\bfz}_\text{sym}).
\end{equation}
As the training pair $(\bfx, y)$ is sampled from the training data, and $\tilde{\bfz}_\text{sym}$ is sampled from $\text{Sel}(\tilde{\bfz}_\text{sym}|\bfx)$, we have that 
\begin{equation}\label{eq:Ezy1}
\begin{aligned}
\mathbb E_{\tilde{\bfz}_\text{sym},y}p(y|\tilde{\bfz}_\text{sym}) = 
\mathbb E_{\bfx,y}p(y|\tilde{\bfz}_\text{sym})p(\tilde{\bfz}_\text{sym}|\bfx)=E_{\bfx,y}p(y|\bfx).\\
\end{aligned}
\end{equation}
We can give each $p(y|\bfx)$ in Eq.~\ref{eq:Ezy1} a $-\log$ to arrive to $-I(\tilde{\bfz}_\text{sym},y)$. Then, it is not difficult to see that 
$-I(\tilde{\bfz}_\text{sym},y)$ has exactly the same form as $L_\text{sp}$.
\end{proof}

\subsection{Derivation of Equation~\ref{eq:mi}}
\label{proof:mi}

Equation~\ref{eq:mi} is the information bottleneck loss for the guider model, this loss encourages that the  features extracted by the guider model are least-but-enough.
\begin{align}
L_\text{mi}=I(\bfx, \bfz_\text{nero})&= \mathbb E_{\bfx, \bfz_\text{nero}}\Big[\log\frac{p(\bfz_\text{nero}|\bfx)}{p(\bfz_\text{nero})}\Big]\\
 &= \mathbb E_{\bfz_\text{nero}}p(\bfx)\Big[\log\frac{p(\bfz_\text{nero}|\bfx)}{p(\bfz_\text{nero})}\Big]\\
&\le \mathbb E_{\bfz_\text{nero}}\Big[\log\frac{p(\bfz_\text{nero}|\bfx)}{p(\bfz_\text{nero})}\Big]\\
&=0.5(\mu^2+\sigma^2-1-2\log\sigma).
\end{align}

\subsection{Proof of Theorem~\ref{the:1}}\label{proof:1}
This theorem is the reason why our language model based regularizer encourages fewer segments of token sequences and decreases bad start or end tokens for each token subsequences, thus makes semantically fluent rationales. The theorem content is restated as follows:
\setcounter{theorem}{0}
 \begin{theorem}
If the following is satisfied for all $i,j$: 
\begin{itemize}
    \item $m'_i<\epsilon \ll 1-\epsilon< m_i$, ($0<\epsilon<1$), and 
    \item $\big|p(m'_ix_i|x_{<i})-p(m'_jx_j|x_{<j})\big|<\epsilon$,
\end{itemize}
then the following two inequalities hold:\\
 (1) $L_\text{lm}(\ldots,m_k,\ldots, m'_{n})<L_\text{lm}(\ldots,m'_k,\ldots, m_{n})$.\\
 (2) $L_\text{lm}(m_1, \ldots,m'_k,\ldots)>L_\text{lm}(m'_1,\ldots,m_k,\ldots)$.
\end{theorem}

\begin{proof}
By Eq.~\ref{eq:lm}, we have:\\
\begin{equation}
\begin{small}
\begin{aligned}
    &L_\text{lm}(\ldots,m'_k,\ldots, m_{n})=-\Big[\sum_{i\neq k,k+1}m_{i-1}\log P(m_ix_i|x_{<i}) + m_{k-1}\log P(m'_kx_k|x_{<k})+m'_k\log P(m_{k+1}x_{k+1}|x_{<k+1})\Big]. 
\end{aligned}
\end{small}
\end{equation}
Therefore, we have the following equation:\\
\begin{equation}\label{eq:pr:1}
\begin{aligned}
    &L_\text{lm}(\ldots,m'_k,\ldots, m_{n})-L_\text{lm}(\ldots,m_k,\ldots, m'_{n})\\
    =&-m_{k-1}\log p(m'_k)+m_{k-1}\log p(m_k)-m'_k\log p(m_{k+1})+m_k\log p(m_{k+1})\\
     &-m_{n-1}\log p(m_n)+m_{n-1}\log p(m'_n)- m_n\log p(m'_{k+1}) + m'_n\log p(m'_{n+1}),\\
\end{aligned}
\end{equation}
where, for simplicity, we use the abbreviation $p(m_k)$ to represent $p(m_kx_k|x_{<k})$.

We also have that:
\begin{align}
- m_n\log p(m'_{k+1})+m_{n-1}\log p(m'_n)&=(m_{n-1}-m_{n})\log p(m'_{k+1}) - m_{n-1}\log\frac{p(m'_{k+1})}{p(m'_n)}\\
&\ge \epsilon\log p(m'_{k+1})-\epsilon
\end{align}
Since ${p(m_k)}_k$ are expected to have large probability values in the language model training process, we have that $p(m_k)>\delta$, and, therefore, $-|\log\delta|<\log\frac{p(m_{k+1})}{p(m_n)}<|\log\delta|$.

Hence, we have that:\\
\begin{align}
- m_{n-1}\log p(m_n)+m_k\log p(m_{k+1})&=(m_k-m_{n-1})\log p(m_{k+1}) + m_{n-1}\log\frac{p(m_{k+1})}{p(m_n)}\\
&\ge \epsilon\log p(m_{k+1})-|\log\delta|\ge(\epsilon-1)|\log\delta|.
\end{align}
Similarly, $-m'_k\log p(m_{k+1})+m_{k-1}\log p(m_k)\ge (1-2\epsilon)\log p(m_k)+m'_k\log\frac{p(m_{k})}{p(m_{k+1})}\ge (1-3\epsilon) |\log\delta|$.\\ 

Therefore, the lower bound of the expression in Eq.~\ref{eq:pr:1} is:\\
\begin{equation}
\begin{aligned}
\inf&=-(1-\epsilon)\log p(m'_k)+ \epsilon\log p(m'_{n+1})+\epsilon\log p(m'_{k+1})-2\epsilon|\log\delta|-\epsilon\\
	&\ge -(1-3\epsilon)\log p(m'_k)-4\epsilon|\log\delta|-\epsilon >0.\\
\end{aligned}
\end{equation}
This proves the statement of the theorem.
\end{proof}

\section{More Results.}

 We list more examples of rationales extracted by our model for the BeerAdvocate dataset in Table~\ref{tab:beermorecase}.

\begin{table}[!h]
\centering
\resizebox{\linewidth}{!}{
\begin{tabular}{|p{10cm}|p{10cm}|}
\hline
\multicolumn{1}{|c|}{Gold} & \multicolumn{1}{c|}{InfoCal}\\
\hline
\textcolor{red}{dark black with nearly no light at all shining through on this one . rich tan colored head of about two inches quickly settled down to about a half inch of tan that thoroughly coated the inside of the glass .} this was what the style is all about \textcolor{green}{the aroma was just loaded down with coffee . rich  notes of mocha mixes in with a rich , and sweet coffee note . a tiny bit of bitterness and an earthy flare lying down underneath of it , but the majority of this one was hands down , rich brewed coffee .} the flavor was more of the same . \textcolor{blue}{rich notes just rolled over the tongue in waves and thoroughly coated the inside of the mouth .} sweet with touches of chocolate and vanilla to highlight the coffee notes &
\textcolor{red}{dark black with nearly no light at all shining through on this one . rich tan colored head of about two inches quickly settled down} to about a half inch of tan that thoroughly coated the inside of the glass . this was what the style is all about \textcolor{green}{the aroma was just loaded down with coffee . rich  notes of mocha mixes in with a rich , and sweet coffee note .} a tiny bit of \textcolor{green}{bitterness and an earthy flare lying down underneath of it }, but the majority of this one was hands down , rich brewed coffee . the flavor was more of the same . \textcolor{blue}{rich notes just rolled over the tongue in waves and thoroughly coated the inside of the mouth} . sweet with touches of chocolate and vanilla to highlight the coffee notes\\ 
\hline
\textcolor{red}{clear copper colored brew , medium cream colored  head . }\textcolor{green}{floral hop nose , caramel malt .} caramel malt front dominated by a nice floral hop backround . grapefruit tones . very tasty hops run the show with this brew .\textcolor{blue}{ thin to medium mouth }. not a bad choice if you 're looking for a nice hop treat .
&
\textcolor{red}{clear copper colored brew , medium cream colored}  head  \textcolor{green}{. floral hop nose , caramel malt .} caramel malt front dominated by a nice floral hop backround . grapefruit tones . very tasty hops run the show with this brew .\textcolor{blue}{ thin to medium mouth }. not a bad choice if you 're looking for a nice hop treat .
\\
\hline
12oz bottle into my pint glass . \textcolor{red}{looks decent , a brown color ( imagine that ! ) with a tan head . nothing bad , nothing extraordinary . }\textcolor{green}{smell is nice , slight roast , some nuttiness , and hint of hops . pretty much to-style . }taste is good but a little underwhelming . toffee malt , some slight roast gives chocolate impressions . hoppiness is mild and earthy . just a touch of bitterness . pretty nondescript overall , but nothing offensive . \textcolor{blue}{mouthfeel is good , medium body and light carb give a creamy finish . }drinkability was nice . i would try this again but wo n't be seeking it out .
&
12oz bottle into my \textcolor{red}{pint} glass . \textcolor{red}{looks decent , a brown color ( imagine that ! ) with a tan head .} nothing bad , nothing extraordinary . \textcolor{green}{smell is nice , slight roast , some nuttiness ,} and hint of hops . pretty much to-style . taste is good but a little underwhelming . toffee malt , some slight roast gives chocolate impressions . hoppiness is mild and earthy . just a touch of bitterness . pretty nondescript overall , but nothing offensive . \textcolor{blue}{mouthfeel is good , medium body and light carb give a creamy} finish . drinkability was nice . i would try this again but wo n't be seeking it out .\\\hline
\end{tabular}
}
\smallskip 
\caption{More instances from the BeerAdvocate dataset. In red the rationales for the appearance aspect, in green the rationales for the smell aspect, and in blue the rationales for the palate aspect. }
\label{tab:beermorecase}
\end{table}%

\begin{figure}[!h]
\begin{center}
\includegraphics[width=\linewidth]{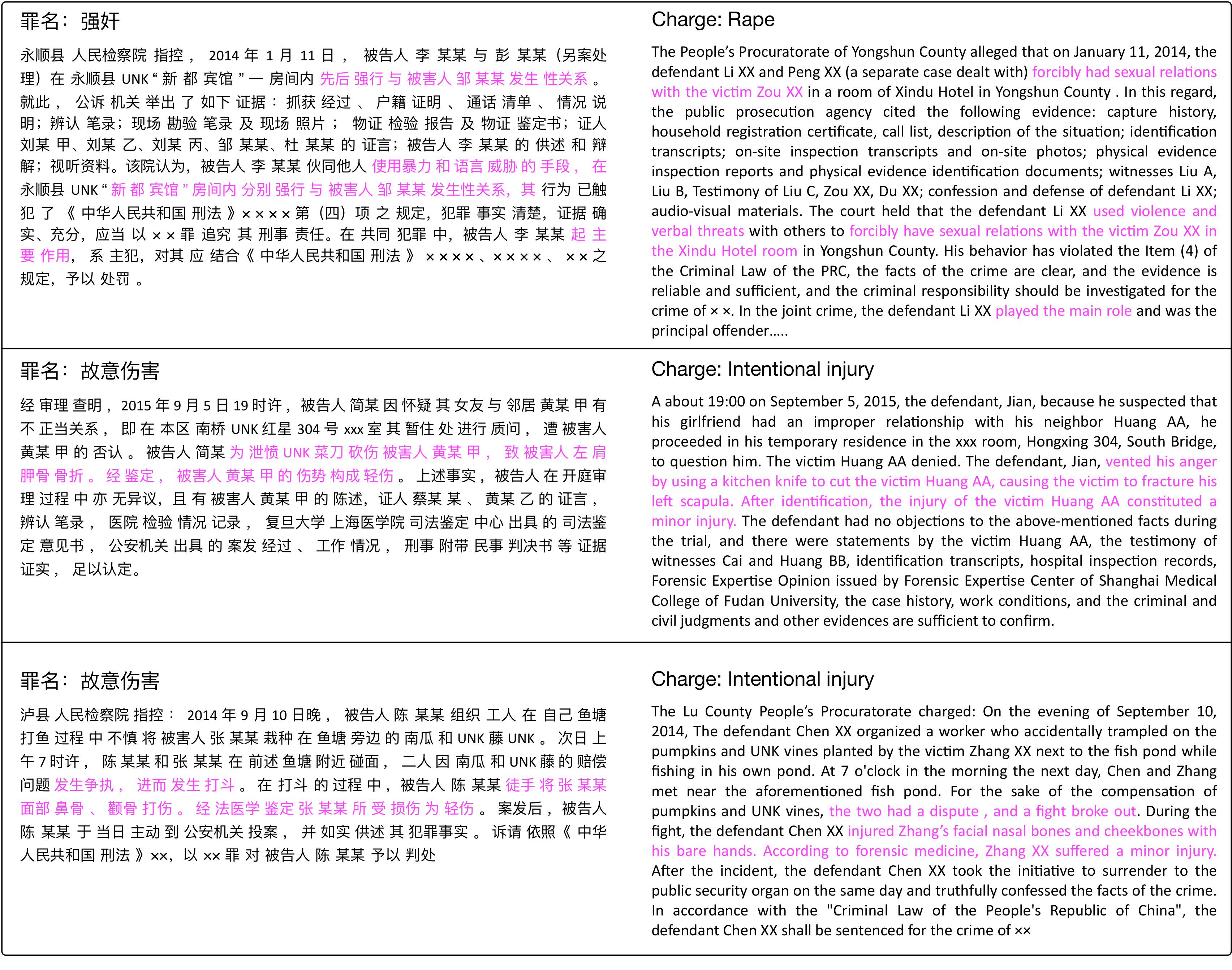}\\
\caption{More instances from the CAIL2018 dataset. Left: the fact description (in Chinese). Right: the corresponding English translation of the fact description. In pink is the selected rationales.}
\label{tab:lawmorecase}
\end{center}
\end{figure}

More examples of  rationales extracted by our model for the legal judgement tasks are shown in Table~\ref{tab:lawmorecase}. 

 \section{Human Evaluation Setup}\label{human}
Our annotators were asked the following questions, in order to assess the usefulness, completeness, and fluency of the rationales provided by our model.

\subsection{Usefulness of Rationales}

Q: Do you think the selected tokens/rationale are \textbf{useful} to explain the ground-truth label?

Please choose a score according to the following description. Note that the score is not necessary an integer, you can give intermediate scores, such as $3.2$ or $4.9$ if you deem appropriate.
\begin{itemize}
\item 5: Exactly. I can give the correct label only by seeing the given tokens.
\item 4: Highly useful. Although most of the selected tokens lead to the correct label, there are still several tokens that have no relation to the correct label.
\item 3: Half of them are useful. About half of the tokens can give some hint for the correct label, the rest are nonsense to the label.
\item 2: Almost useless. Almost all of the tokens are useless, but there are still several tokens that are useful. 
\item 1: No Use. I feel very confused about the selected tokens, I don't know which law article/charge/term of penalty the article belongs to.
\end{itemize}

\subsection{Completeness of Rationales}

Q: Do you think the selected tokens/rationale are \textbf{enough} to explain the ground-truth label?

Please choose a score according to the following description. Note that the score is not necessary an integer, you can give intermediate scores, such as $3.2$ or $4.9$, if you deem appropriate.
\begin{itemize}
\item 5: Exactly. I can give the correct label only by the given tokens.
\item 4: Highly complete. There are still several tokens in the fact description that have a relation to the correct label, but they are not selected.
\item 3: Half complete. There are still important tokens in the fact description, and they are in nearly the same number as the selected tokens.
\item 2: Somewhat complete. The selected tokens are not enough. There are still many important tokens in the fact description not being selected. 
\item 1: Nonsense. All of the selected tokens are useless. None of the important tokens is selected.

\end{itemize}
\subsection{Fluency}

Q: How fluent do you think the selected rationale is?  For example: \textit{``He stole an iPhone in the room''} is very fluent, which should have a high score. \textit{``stole iPhone room''} is just separated tokens, which should have a low fluency score.

Please choose a score according to the following description. Note that the score is not necessary an integer, you can give scores like $3.2$ or $4.9$ , if you deem appropriate.
\begin{itemize}
\item 5: Very fluent. 
\item 4: Highly fluent. 
\item 3: Partial fluent. 
\item 2: Very unfluent. 
\item 1: Nonsense.
\end{itemize}


\clearpage
\bio{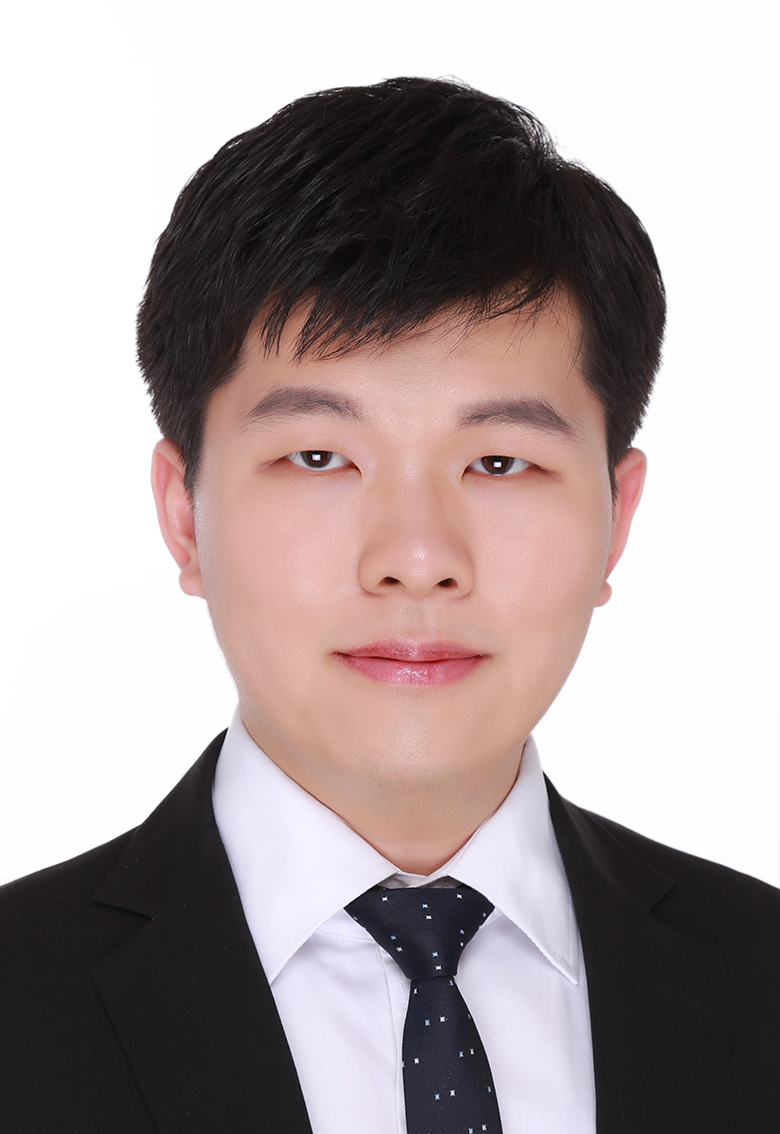}
\textbf{Lei Sha} is currently  a tenure-track associate professor at Beihang University. He was a research associate at the Department of Computer Science, University of Oxford. He was advised by Prof. Thomas Lukasiewicz in the Intelligent System Lab. Previously, he was an NLP research scientist in Apple. While at Apple, he was responsible for Siri’s module, such as domain classification and chit-chat dialogue. Before that, in 2018, he obtained his PhD degree and graduated from Peking University, China. During his PhD period, he focused on the research of learning and generating from structured data. He also served as a research assistant in Microsoft Research Asia, working with Chin-yew Lin, Lintao Zhang, and Qi Chen. 
He has published many cutting-edge research papers in event extraction, text entailment recognition, and text generation. These first-author papers are published in top conferences of NLP, such as ACL, EMNLP, NAACL, AAAI, etc. Also, he was the senior program committee of IJCAI 2021, and the reviewer of many top-tier conferences and journals, such as AAAI, ACL, TASLP, EMNLP, IJCNLP, etc. He also achieved many top awards, such as Lee-Wai Wing Scholarship and May 4th Scholarship, which is the top scholarship of Peking University.  His research interest focuses on natural language understanding and controllable text generation
\endbio

\bio{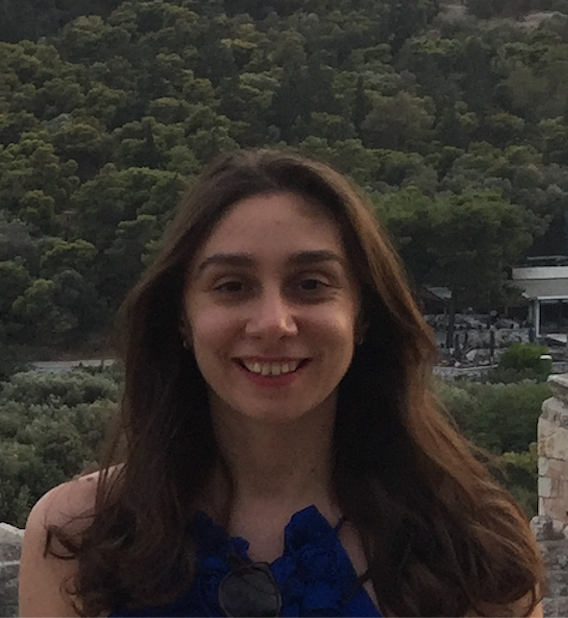}
\textbf{Oana-Maria Camburu} is currently a Research Associate at the Department of Computer Science at University of Oxford and a co-investigator at the Alan Turing Institute on the project ``Neural Networks with Natural Language Explanations''. Oana has done her undergraduate and MSc studies at the Ecole Polytechnique, Paris, with a focus on Applied Mathematics and Machine Learning. She obtained her PhD from the Department of Computer Science at University of Oxford, on the topic of explainable AI. Her excellent dissertation has been nominated for the ACM Doctoral Dissertation Award 2020 (maximum 2 nominations per university among the topics of computing and engineering) as well as at the Joint AAAI/ACM SIGAI Doctoral Dissertation Award 2020 (maximum 1 nomination per university on AI topics; both currently ongoing competitions). She also obtained a J.P. Morgan PhD Fellowship on the topic of explainability.
She published in top-tier venues such as NeurIPS, ACL, EMNLP, CVPR, AAAI.
She is currently an organizer for two workshops: ``The 6th Workshop on Representation Learning for NLP'' (RepL4NLP-2021) – accepted at ACL 2021, and 
``Towards Explainable and Trustworthy Autonomous Physical Systems'' – accepted at ACM CHI 2021.
Previously, she undertook a series of internships at prestigious companies and research centers, such as Google (2 internships, in London and Mountain View), the Operational Research Center at Massachusetts Institute of Technology (MIT), Cambridge, Massachusetts, and Max Planck Institute for Mathematics, Bonn, Germany.
\endbio

\bio{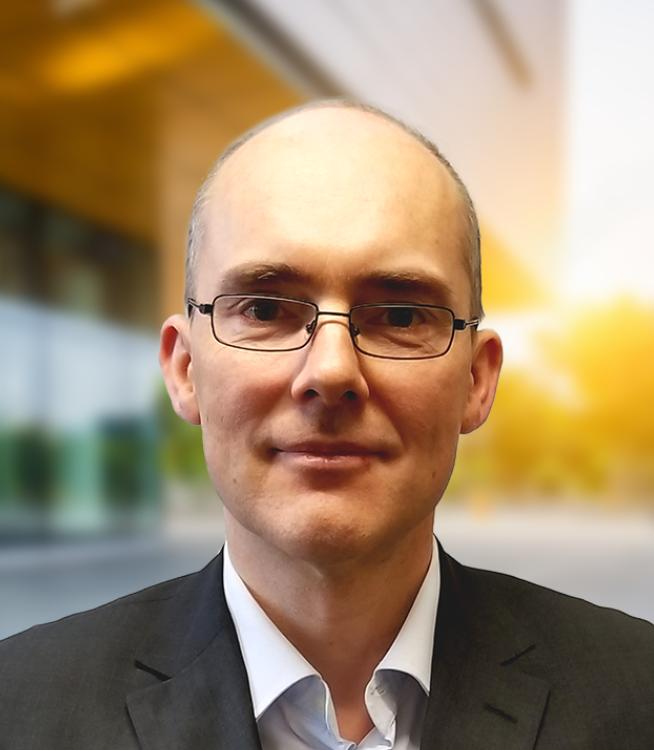}
\textbf{Thomas Lukasiewicz} is a Professor of Computer Science and
Yahoo!\ Research Fellow at OUCS since 2010
and a Faculty Fellow at the Alan Turing Institute since 2016. Prior to this, from 2004 to 2009, he was holding a prestigious
Heisenberg Fellowship by the German Research Foundation (DFG), affiliated with OUCS, the Institute of Information Systems, TU Vienna, Austria,
and the Department of Computer and System Sciences, Sapienza University of Rome, Italy.
His research interests are in information systems and artificial intelligence (AI), including especially
knowledge representation and reasoning, uncertainty in AI, machine learning, the (Social and/or Semantic) Web, and data\-bases.
He has published more than 200
publications, many of them at top-tier and leading international conferences and journals, 
including many highly cited papers. He received the IJCAI-01 Distinguished Paper Award (for the
best paper at IJCAI-01), 
the AIJ Prominent Paper Award 2013 (for the best paper in the journal AIJ between 2008 and 2013), and the RuleML~2015 Best Paper Award.
He 
has been a PC member of more than 150 conferences and
workshops (more than 20 of which \mbox{(co-)}chaired). He has given invited talks and
invited tutorials at many conferences and workshops. He 
is area editor for  ACM TOCL,
associate editor for JAIR and AIJ,
and editor 
for {Semantic Web}. 
He has acted as principal investigator for many successfully completed grants funded
by the DFG, the Austrian Science Fund, the EU,
the EPSRC, and Google. He is currently 
the principal investigator for a seed funding grant at the Alan Turing Institute 
and  for an EPSRC Doctoral Prize, and 
a co-investigator for the DBOnto and VADA EPSRC platform and programme grants, respectively.
\endbio
\end{document}